\documentclass[10pt]{article}
\usepackage[preprint]{tmlr}

\usepackage{amsmath}
\usepackage{amssymb}
\usepackage{graphicx}
\usepackage{booktabs}
\usepackage{float}
\usepackage{hyperref}
\usepackage{url}
\usepackage{natbib}
\usepackage{amsthm}

\newtheorem{proposition}{Proposition}

\title{TrustMix: Graceful Degradation for\\ Pretrained Models under Distribution Shift}

\author{\name Juliusz Ziomek \email juliusz [at] robots (dot) ac.uk\\
       \addr Machine Learning Research Group\\
       University of Oxford\\
       Oxford OX1 2JD, United Kingdom
       \AND
       \name Maike Osborne \email maike.osborne [at] eng (dot) ox.ac.uk \\
       \addr Machine Learning Research Group\\
       University of Oxford\\
       Oxford OX1 2JD, United Kingdom
       \AND
       \name Yee Whye Teh \email yee.teh [at] stats (dot) ox.ac.uk \\
       \addr Department of Statistics\\
       University of Oxford\\
       Oxford OX1 2JD, United Kingdom}

\def\month{MM}
\def\year{YYYY}
\def\openreview{\url{https://openreview.net/forum?id=XXXX}}

\begin{document}

\title{TrustMix: Graceful Degradation for\\ Pretrained Models under Distribution Shift}

\author{\name Juliusz Ziomek \email juliusz [at] robots (dot) ac.uk\\
       \addr Machine Learning Research Group\\
       University of Oxford\\
       Oxford OX1 2JD, United Kingdom
       \AND
       \name Maike Osborne \email maike.osborne [at] eng (dot) ox.ac.uk \\
       \addr Machine Learning Research Group\\
       University of Oxford\\
       Oxford OX1 2JD, United Kingdom
       \AND
       \name Yee Whye Teh \email yee.teh [at] stats (dot) ox.ac.uk \\
       \addr Department of Statistics\\
       University of Oxford\\
       Oxford OX1 2JD, United Kingdom}

\maketitle

\begin{abstract}
Predicting beyond the training domain necessarily requires assumptions about how the ground truth extrapolates, yet in most neural networks these assumptions are made only implicit through the architecture. We propose TrustMix, a lightweight post-hoc method that makes a pretrained model’s extrapolative behaviour explicit and controllable. TrustMix  combines the predictive distribution produced by a pretrained model with a user-specified fallback distribution $p_0(y|x)$, such that as the out-of-distribution (OOD) deviation statistic grows, we collapse towards $p_0(y|x)$.  TrustMix provides graceful degradation: the  predictive distribution converges toward the fallback distribution as the OOD signal grows. The fallback distribution $p_0(y|x)$ can reflect domain knowledge, if available, or rely on  a domain-agnostic stationarity assumption. TrustMix does not require retraining of the base model and can be applied to both regression and classification problems. We evaluated TrustMix on several synthetic and real-world distribution shifts, including tabular, image, and natural language processing tasks. We tested TrustMix on datasets such as CIFAR-10/100 and ImageNet and on numerous architectures including Convolutional Neural Networks (CNNs) and Transformers. TrustMix remains competitive in-distribution while improving predictive uncertainty under distribution shift, with gains in predictive likelihood for regression and calibration for classification. Finally, we showed that TrustMix can provide useful predictive distributions for decision-making applications by utilizing it as a surrogate for Bayesian Optimisation.
\end{abstract}


\section{Introduction}

Nothing is more dangerous than a failure that is silent. Neural networks today are capable of achieving excellent performance within training distribution. However, their behaviour outside of this training domain is less understood. A network can be confident even when it is completely wrong and far from the domain it was trained upon. In general, making predictions under distribution shift requires assumptions about how patterns observed in the training data transfer beyond the training domain. Standard neural networks make such assumptions implicitly through their architecture; however, they lack a mechanism directly identifying the transitions from interpolation to extrapolation. To address this, Out-Of-Distribution (OOD) Detection methods have been developed.

Although OOD detection signals can identify inputs that appear to be unusual or atypical, they do not resolve the fundamental problem of prediction. An OOD score can indicate that the relationship between inputs and outputs that a model has learned may no longer apply; however it does not specify what the model should predict instead.  While the optimal predictive distribution in those regions necessarily depends on domain assumptions, we believe that these assumptions can be made explicit. We present \textsc{TrustMix}, a simple post-hoc framework for enabling graceful degradation of pretrained models during distribution shift. TrustMix separates the predictive behaviour that a model learns from data and the assumed fallback distribution $p_0(y \mid x)$ to which the model collapses in OOD regimes. TrustMix gradually transitions from the base model's predictive distribution to the fallback distribution, as a scalar deviation signal $D(x)$ grows. We specifically enforce the monotonicity of this transition architecturally, which guarantees that reliance on the base model is non-increasing as a function of $D(x)$ and disappears entirely as $D(x) \rightarrow \infty$.  TrustMix does not eliminate model assumptions, but makes them controllable and explicit through the choice of fallback distribution $p_0(y \mid x)$.  \paragraph{Contributions.} \begin{itemize} \item We introduce \textbf{TrustMix}, a straightforward post-hoc mechanism that endows a pretrained neural network with \emph{graceful degradation}: as inputs move increasingly far from the training distribution, predictions made by TrustMix will converge to a user-defined fallback distribution without  modifying the original base model. \item We enforce that reliance on the base model decreases monotonically with deviation and vanishes asymptotically (Proposition 1), for any parameter values. 
\item We demonstrate TrustMix across regression and classification tasks, multiple model architectures, and both synthetic and real-world distribution shifts. We further show that its predictive distributions are useful for downstream decision-making by applying TrustMix as a scalable neural surrogate for Bayesian optimisation.\end{itemize}

\section{Problem Statement and Preliminaries} \label{sec:prelim}

We assume access to a pretrained base model consisting of a feature extractor
$\Phi: \mathcal{X} \rightarrow \mathbb{R}^{d_z}$ and a task-specific prediction head
$h: \mathbb{R}^{d_z} \rightarrow \Delta(\mathcal{Y})$, where $\Delta(\mathcal{Y})$ denotes the set of distributions over $\mathcal{Y}$. The resulting predictive distribution is obtained as
$f(x) = h(\Phi(x))$.
We assume the model is trained on a dataset $\mathcal{D}_{\mathrm{train}}$ and we have access to some validation set $\mathcal{D}_{\mathrm{val}}$, unseen by the model during training.
Throughout, $\Phi(\cdot)$ and $h(\cdot)$ are fixed and our goal is to endow the model with graceful degradation under distribution shifts, without modifying the base model itself. Given a proper scoring rule $S(q, y)$ measuring quality of predictive distribution $q(y|x)$ when the outcome is $y$, the risk under the true distribution $p^\star(y|x)$ at point $x$ is given by $R(q, x) = \mathbb{E}_{y \sim p^\star(y|x)}[-S(q,y)]$. We thus wish to produce an output distribution with the lowest expected score. Throughout, we rely on log-likelihood $S(q,y) = \log q(y|x)$ as the scoring rule, but note that other scoring rules can also be used.

We assume we have access to some OOD deviation signal $D(x): \mathcal{X} \to \mathbb{R}^+$ produced by the model, which quantifies how far the input is from the training distribution. However, even a perfect OOD signal does not directly specify how predictions should be adjusted: it is a single scalar and does not directly map to a predictive distribution. A plethora of OOD signals have been proposed in past literature, which we describe in detail in Section \ref{sec:related_work}. Throughout the paper we rely on Mahalanobis distance as a robust default, as it is known for strong OOD detection performance \citep{mukhoti2023deep}, and importantly can be both applied to both regression and classification. In Appendix \ref{app:mah} we detail how the Mahalanobis distance can be efficiently computed. Alternative signals such as the energy score \citep{liu2020energy} are also applicable, which we investigate in Section \ref{sec:addexp}.

Crucially, even a perfect deviation signal does not resolve the
problem of making predictions under distribution shift. The signal $D(x)$ can provide evidence about
whether the relationship learned by the base model remains applicable at
$x$, but it does not specify what the predictive distribution should be when
that relationship is no longer trusted. We therefore separate two questions: \emph{when should the base model be
trusted}, and \emph{what should we predict when it is not?} TrustMix uses
$D(x)$ to address the former and an explicit fallback distribution to encode
the modelling assumption required for the latter. The usefulness of $D(x)$
ultimately depends on whether it is informative about predictive failure. In
Section~\ref{sec:addexp}, we empirically investigate this assumption and show
that larger representation-space deviations are associated with deteriorating
base-model predictive performance across the considered tasks.

\section{Methodology}

When a model's inputs diverge from the training distribution, its
predictions need not become visibly unreliable - there is no guarantee their uncertainty will increase. Previous literature shows that a fully connected network with ReLU non-linearities tend to simply continue the linear trend, eventually exploding to $\pm \infty$ \citep{ziomek2025just}, thus making very extreme predictions in regions where we should rather be cautious.  As established in Section~\ref{sec:prelim}, an
OOD deviation signal $D(x)$ on its own only quantifies how atypical an
input is; it does not prescribe how a model's prediction should change in
response. Our goal is therefore to equip a fixed, pretrained model
$f = h \circ \Phi$ with the capacity to gracefully reduce its reliance on
its own prediction as the input $x$ moves away from the training
distribution, without retraining $\Phi$ or $h$.
Since predictive behaviour far from the training distribution cannot in
general be determined from training data alone, we explicitly specify the predictive
behaviour we wish the model to approach when its own prediction is no longer
trusted.
We propose to smoothly interpolate between two distributions: the model distribution $p_f(y|x)$ and a fallback distribution $p_0(y|x)$, that is:
 \begin{align} \label{eq:trustmix_predictive}
     p(y|x) &= p_f(y|x)\mathcal{T}_{\psi}(x) + p_0(y|x)(1-\mathcal{T}_{\psi}(x)),
 \end{align}
 where $\mathcal{T}_{\psi}(x) $ is the weight in $[0,1]$ given by some gating mechanism, parametrised by $\psi$. We would like to have the property that as $D(x) \to \infty$,  the gating weight $\mathcal{T}_{\psi}(x) \to 0$ and the predictive distribution collapses to the fallback $p_0(y|x)$. As such, we use:
 \begin{align}
     \mathcal{T}_{\psi}(x) = \text{sigmoid}(-g(D(x);\psi))
 \end{align}
where $g(\cdot;\psi)$ is a small, monotonic neural network parametrised with $\psi$, which is guaranteed to be monotonically non-decreasing in its input. This can be easily achieved by ensuring all its weights are positive and using non-decreasing activations, as stated by the next proposition.
\begin{proposition}[Architectural monotonicity]
\label{prop:monotone}
Let $g(D) = W_L\,\phi(W_{L-1}\,\phi(\cdots\phi(W_1 D + b_1)\cdots) +
b_{L-1}) + b_L$, where every $W_l$ has positive entries and $\phi$ is
non-decreasing and unbounded. Then $g(\cdot)$ is non-decreasing in $D$, and
hence $\text{sigmoid}(-g(D(x);\psi))$ is
non-increasing in $D(x)$ and as
$D(x)\to\infty$, we get $\mathcal{T}(x)\to 0$.
\end{proposition}
\begin{proof}
Appendix~\ref{ap:monotone_proof}.
\end{proof}
The conditions of the proposition above hold if we use the reparametrisation $W_l = |\omega_l| + \mathcal{I} \mathcal{I}^T\epsilon$ for some small $\epsilon > 0$ and optimise parameters $\omega_1,\dots,\omega_L$ instead. Across all experiments we use the same architecture with two hidden layers, each with 16 neurons and  batch normalisation between the hidden layers. For batch normalisation, we turn off the learnable affine parameters, so that it only centres and rescales its input by a positive standard deviation, thus preserving non-decreasing and limiting properties of $g(D)$. This guarantees collapse to the fallback distribution $p_0$ for extreme $D$. We now discuss how this fallback can be chosen.

\paragraph{Fallback Distribution}
The fallback distribution $p_0(y|x)$ represents the predictive behaviour we
wish to adopt when we completely stop trusting the base model. Importantly,
TrustMix does not assume that this distribution can be inferred from the
training data: sufficiently far from the observed domain, predictive behaviour
necessarily depends on additional modelling assumptions. The fallback makes
these assumptions explicit. The behaviour of TrustMix under extreme distribution shifts thus depends on the choice of $p_0$; TrustMix governs the transition behaviour, but does not  guarantee that arbitrarily chosen $p_0$ is well specified.
When relevant domain knowledge is available, it
can be incorporated directly into $p_0(y|x)$. In the absence of such
information, we use a simple domain-agnostic assumption that the global moments
of the predictive distribution remain stationary throughout the domain. Let us
define:
\begin{align}
\label{eq:fallback_params}
    \mu_0
    =
    \frac{1}{|D_{\mathrm{train}}|}
    \sum_{(x,y)\in D_{\mathrm{train}}} f(x),
    \qquad
    \sigma_0^2
    =
    \frac{1}{|D_{\mathrm{train}}|}
    \sum_{(x,y)\in D_{\mathrm{train}}}
    \bigl(f(x)-\mu_0\bigr)^2,
\end{align}
to be the first and second central moments of model predictions at training points. In the regression setting, we propose to simply use:
\begin{align}
    p_0(y \mid x) &= \mathcal{N}\!\big(y;\mu_0, \sigma_0^2\big) \label{eq:fallback},
\end{align}
as the default, input-independent fallback distribution. In Section \ref{sec:addexp}, we investigate how accurate the fallback models observation under extreme deviation relatively to the base model. Note that such an assumption is common in regression models. For example, Gaussian Process (GP) regressors typically collapse to the normal prior with fitted mean and scale as the input deviates from training data \citep{rasmussen2006gaussian}. In classification, we put the same Gaussian fallback, except on the model logits, as we explain later in this section. 
We now proceed to discuss details of applying TrustMix in regression and classification settings.

\begin{figure}[t]
    \centering
    \includegraphics[width=\linewidth]{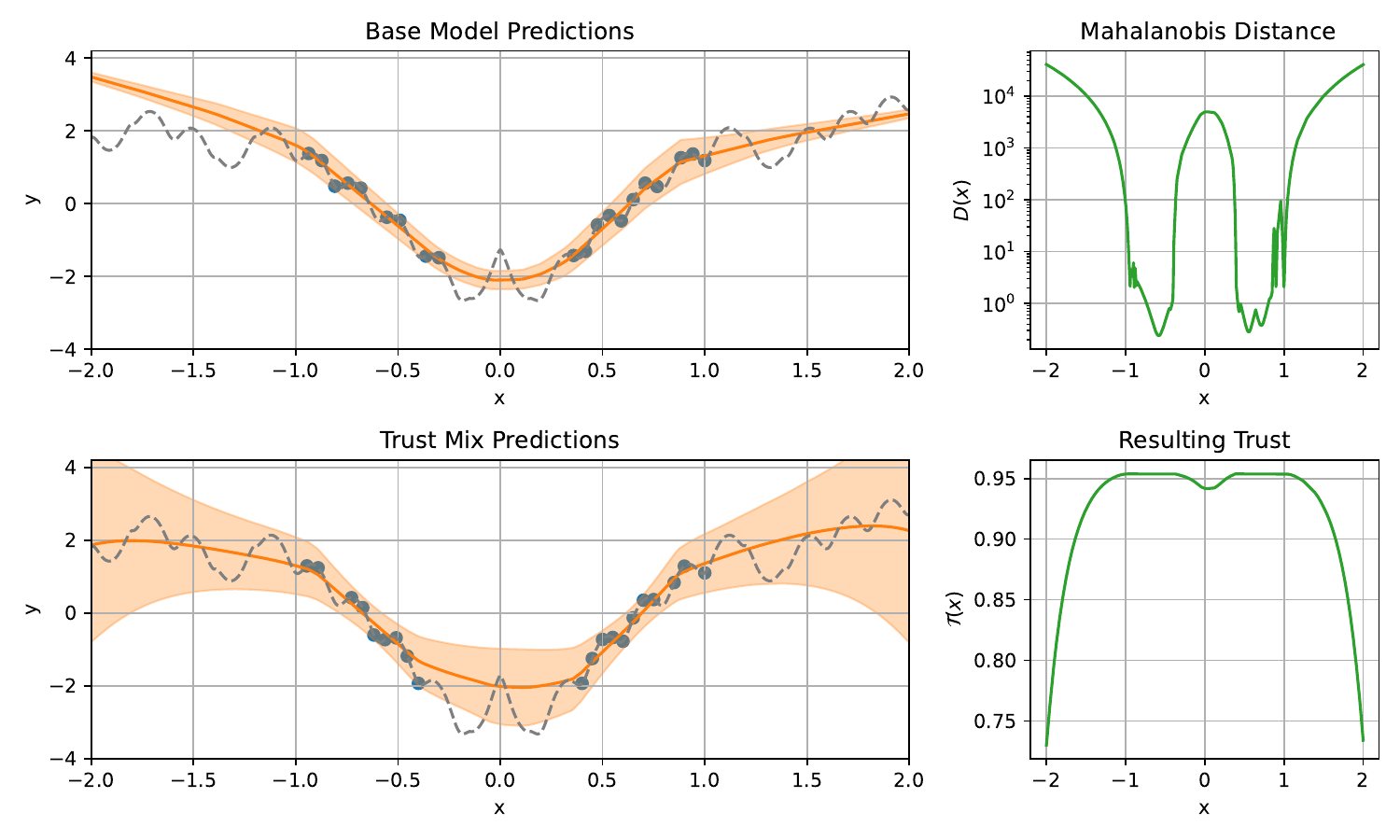}
     \caption{
Illustration of TrustMix on a toy regression problem.
Top-left: predictions of a base MLP with a variance head, showing overconfident extrapolation outside the training region.
Top-right: Mahalanobis deviation score $D(x)$, which increases sharply away from the training distribution.
Bottom-right: resulting mixture weight $\mathcal{T}_\psi(x)$, which decreases with deviation.
Bottom-left: TrustMix predictions obtained by mixing the base predictor with the fallback distribution, producing increased uncertainty under distribution shift.
}
    \label{fig:toyproblem}
\end{figure}
 \paragraph{TrustMix for Regression} If the model is equipped with a variance head $\hat{\sigma^2_f}(x)$, we simply set $\sigma_f(x;\rho) = \rho \hat{\sigma}_f(x)$ for some learnable $\rho > 0$. Otherwise we use a constant variance $\sigma_f(x;\rho) = \rho$. We include a variance scaling parameter to correct systematic over- or under-confidence in the base predictive uncertainty estimates.
\begin{align}
p_f(y \mid x)  &= \mathcal{N}\!\big(y;f(x), \sigma^2_f(x; \rho)\big) \nonumber
\end{align}
The overall predictive distribution from Equation \ref{eq:trustmix_predictive} has the following mean and variance:
\begin{align*}
\mu(x;\psi) &= \mathcal{T}_\psi(x) f(x) + (1-\mathcal{T}_\psi(x)) \mu_0, \\
\sigma^2(x; \psi, \rho) &= \mathcal{T}_\psi(x)\sigma^2_f(x;\rho) + (1-\mathcal{T}_\psi(x))\sigma_0^2  + \mathcal{T}_\psi(x)(1-\mathcal{T}_\psi(x))(f(x) - \mu_0)^2.
\end{align*}
 For computational convenience in regression, we approximate this mixture by a Gaussian whose first two moments match those of the mixture:
\begin{align*}
    p(y \mid x;\psi, \rho) = \mathcal{N}( y;\mu(x;\psi), \sigma^2(x;\psi, \rho)).
\end{align*}

\paragraph{TrustMix for Classification.}
For classification, we apply TrustMix in logit space rather than directly
to class probabilities. We first transform the base model logits using
an affine transformation $\alpha f(x) + \beta$, where
$\alpha \in \mathbb{R}^{+}$ and $\beta \in \mathbb{R}$ are learnable
calibration parameters. This allows us to adjust the scale and offset
of the logits before applying TrustMix.

We then model the logits using a diagonal Gaussian distribution, with
mean $\mu(x;\theta) \in \mathbb{R}^C$ and variance
$\sigma^2(x;\theta) \in \mathbb{R}^C$, where
$\theta = (\psi,\rho,\alpha,\beta)$. The fallback distribution is also
defined as a diagonal Gaussian in logit space, with its mean and
variance estimated from the base model's predictions on the training
data. TrustMix interpolates between the transformed base logits and
this fallback distribution, with the interpolation controlled by the
learned trust function.

To obtain class probabilities, we use the analytical approximation
of \citet{mucsanyi2025rethinking}:
\begin{equation*}
    p(y \mid x;\theta)
    =
    \operatorname{softmax}\left(
        \frac{\mu(x;\theta)}
        {\sqrt{1+\frac{\pi^2}{8}\sigma^2(x;\theta)}}
    \right),
\end{equation*}
where division and the square root are applied element-wise.
Under this approximation, the variance of each logit attenuates
its contribution to the softmax independently. Consequently,
changes in logit uncertainty can affect both the sharpness of
the predictive distribution and the relative probabilities
of different classes.

An alternative would be to interpolate directly between the base
model's class probabilities and a fallback categorical distribution.
Such an approach restricts the resulting predictions to convex
combinations of these two probability vectors. In contrast,
TrustMix performs the interpolation in logit space and incorporates
the resulting uncertainty before applying the nonlinear softmax
transformation. This allows the predictive distribution to change
in ways that are not restricted to linear interpolation between
two fixed categorical distributions. Further details of the
classification formulation are provided in
Appendix~\ref{ap:tm_class}.

\paragraph{Training TrustMix}
We optimise the following set of parameters $\theta=(\psi, \rho, \alpha, \beta)$. To do so, we optimise the negative log-likelihood over the validation dataset $\mathcal{D}_{\textrm{val}}$ that the base model \textbf{did not see} during training. We thus solve the following:
\[
\max_{\theta} \sum_{(x,y) \in \mathcal{D}_{\textrm{val}}} S(p(\cdot \mid x_i, \theta), y_i) = \min_{\theta} \sum_{(x,y) \in \mathcal{D}_{\textrm{val}}} -\log p(y_i \mid x_i, \theta).
\]
 This objective calibrates the trust weight, encouraging lower reliance on the base model in regions where its predictions are less reliable.  TrustMix can increase the predictive variance  by increasing the rate at which $\mathcal{T}_\psi(x)$ collapses to zero, but doing so will necessarily increase predictive variance for all points where $D(x)$ is large. As such, the only way for TrustMix to decrease the loss above is by appropriately calibrating the relationship between $D(x)$ and the mixing weight $\mathcal{T}_\psi(x)$. Importantly, this calibration requires no out-of-distribution examples: TrustMix learns when to distrust the base model solely from in-distribution validation data. This raises a natural question: can the  parameters be identified from in-distribution data alone? Even within the in-distribution validation dataset, datapoints can exhibit different levels of deviations, allowing TrustMix to leverage the difference in $D(x)$ to calibrate its collapse mechanism. We investigated this mechanism further in Section \ref{sec:addexp}.

\section{Related Work} \label{sec:related_work}

\paragraph{Uncertainty estimation.}
A large body of work aims to quantify predictive uncertainty in deep models, including Bayesian neural networks \citep{neal2012bayesian, blundell2015weight}, MC dropout \citep{gal2016dropout}, deep ensembles \citep{lakshminarayanan2017simple}, and kernel-based approaches such as DUE, DUQ, and SNGP \citep{van2021feature, van2020uncertainty, liu2020simple}. These methods typically require modifying or retraining the base model. In contrast, TrustMix is a post-hoc approach that operates on a fixed pretrained model. These directions are complementary: improved uncertainty estimates from such models can be further calibrated using TrustMix.

\paragraph{Post-hoc calibration.}
Post-hoc calibration methods align predicted probabilities with empirical accuracy without retraining. Temperature scaling \citep{guo2017calibration} and its variants adjust prediction sharpness, while isotonic regression \citep{zadrozny2002transforming} and order-preserving transformations \citep{rahimi2020intra} learn more flexible mappings. However, these methods are primarily designed for in-distribution calibration and do not explicitly account for distributional shift.

Density-aware calibration (DAC) \citep{tomani2023beyond} extends this setting by incorporating representation-space density and inflating temperature as we move to regions with lower estimated density. In contrast, TrustMix learns a simple monotone mapping from a deviation signal to predictive trust.

\paragraph{Regression calibration.}
Calibration for regression is less studied. ~\cite{laves2020well} propose variance scaling, which rescales predictive uncertainty by a global factor. While effective, such methods are input-independent, as they globally scale variance by the same scalar across the domain. TrustMix instead introduces an input-dependent mechanism, allowing predictive uncertainty to adapt to distributional deviation.

\paragraph{Out-of-distribution detection.}
OOD detection methods \citep{hendrycks2016baseline, lee2018simple, nalisnick2019detecting, liu2020energy} assign scores that measure how atypical an input is relative to the training distribution. These methods focus on ranking inputs rather than producing calibrated predictions. TrustMix complements OOD detection by mapping deviation scores to predictive distributions, enabling evaluation under proper scoring rules.

\paragraph{Conformal prediction.}
Conformal prediction \citep{vovk2005algorithmic, angelopoulos2021gentle} provides distribution-free uncertainty quantification via set-valued predictions with coverage guarantees. Unlike TrustMix, it does not produce full predictive distributions and typically assumes exchangeability, which may not hold under distribution shift. The two approaches are orthogonal: conformal methods provide coverage guarantees, while TrustMix enables calibrated probabilistic predictions under shift.

\section{Experiments}

We evaluate TrustMix on regression and classification tasks across diverse datasets and architectures. In all settings, we first train a base model and then apply post-hoc calibration on a held-out validation set. For regression, we use a standard base model that predicts a mean and variance (``VarHead''), trained by maximising Gaussian likelihood, as well as a version with MC Dropout. We compare TrustMix against $\sigma$-scaling \citep{laves2020well} calibration applied on top of both the base model and MC Dropout. Additionally, since none of the other baselines relies on the deviation signal, we also compare against a modified version of $\sigma$-scaling, which we call Distance Aware $\sigma$-scaling, where we scale the base-model predictive standard deviation $\sigma(x)$ as:
\begin{equation*}
    (A  + B \, D(x))\sigma(x),
\end{equation*}
where $A>0$ and $B>0$ are scalars learnt on calibration data and $D(x)$ is the exact same  deviation signal used by TrustMix. Note that $B=0$ recovers standard $\sigma$-scaling.
For classification, we calibrate pretrained models using TrustMix and standard baselines including temperature scaling (TS) \citep{guo2017calibration}, its entropy variant (HTS) \citep{balanya2024adaptive}, isotonic regression \citep{zhang2020mix}, and intra order-preserving networks (IOP-IO) \citep{rahimi2020intra}, as well as density-aware calibration (DAC)\citep{tomani2023beyond}. Further details are provided in Appendix~\ref{ap:expdetails}. Throughout these experiments, we use the Mahalanobis distance as the base OOD deviation signal $D(x)$; details of its computation are provided in Appendix \ref{app:mah}.

\begin{figure}
    \centering
    \includegraphics[width=\linewidth, trim={0 0 0 1cm},clip]{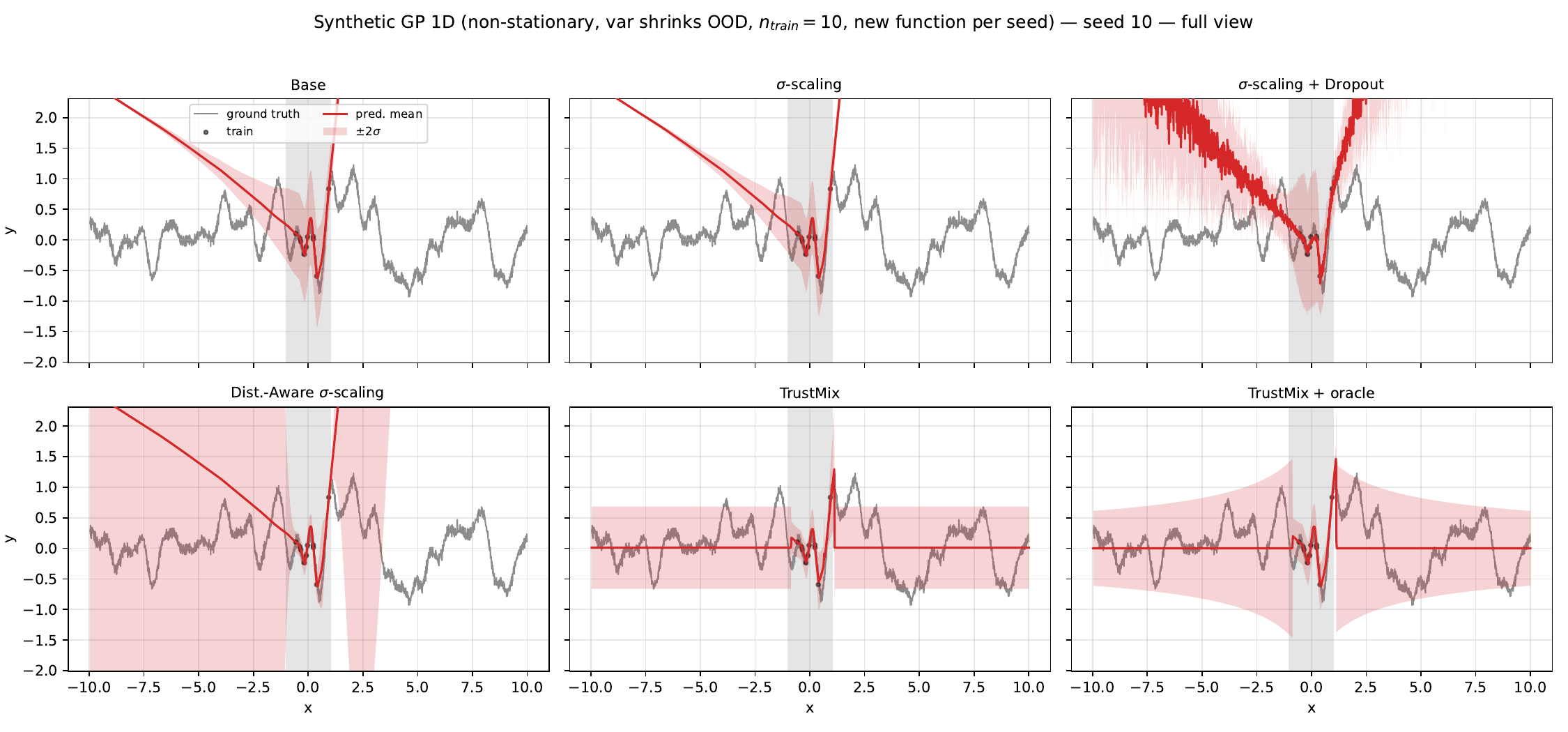}
    \caption{Predictive fit on the 1‑D synthetic GP (non‑stationary kernel, prior variance shrinking out‑of‑distribution). Base and $\sigma$‑scaling extrapolate a confident trend and diverge; MC‑Dropout is erratic; TrustMix reverts the mean to its fitted fallback with a flat uncertainty band; TrustMix + oracle reverts to the true prior.}
    \label{fig:syntheticgp}
\end{figure}

\subsection{Synthetic Experiments}
We start by applying TrustMix to a synthetic dataset where the function is sampled from a noisy (with noise std. deviation $\sigma_n=0.05$) GP prior over a ten-dimensional ($d=10$)  space. 1000 training and 200 validation points  are drawn uniformly from $[-1,1]^d$; whereas evaluation is done on 2000 points drawn from an extended domain of $[-2,2]^d$. We use an isotropic RBF kernel $k^{\text{RBF}}$ \citep{rasmussen2006gaussian} with lengthscale  $\ell = 0.3\sqrt{d}$ (i.e. $\approx 0.95$ for $d=10$) and
 an input-dependent scale, that is:
\begin{equation*}
    k(x,x') = c(x)c(x')k^{\text{RBF}}(x,x').
\end{equation*}
We consider two cases: stationary where $c(x) = 1$ and non-stationary where $c(x) = 1 / \sqrt{(1 + \lVert x\rVert_2)}$.  Intuitively, in the second case, the average function magnitude we expect to see far from training data is actually lower than in-distribution. Importantly, note that before any data is observed, at any point the distribution of observations is $p(y|x) = \mathcal{N}(y;0, k(x,x) + \sigma^2_n)$. We thus consider two variants: TrustMix using the default fallback distribution from Equation \ref{eq:fallback} and a TrustMix with an oracle fallback using $p_0(y|x) = \mathcal{N}(y;0, k(x,x) + \sigma^2_n)$. We present results in Table \ref{tab:synthgp_nll_10d}. In Figure \ref{fig:syntheticgp}, we show example model fits for different methods in the one dimensional case with the non-stationary kernel. In the table, we also include the NLL under the true GP posterior fitted on training data, as the theoretical lower bound. Unsurprisingly, TrustMix with the oracle fallback distribution performs best, followed by the TrustMix with the default fallback given by Equation \ref{eq:fallback}. The remaining methods exhibit substantially higher NLL. The gap between using the oracle and default fallback is particularly pronounced in the non-stationary case, which is expected as in the stationary case, the marginal target distribution far from training data is the same as on the training data. As such, the default fallback, which simply estimates the first and second moments of model predictions on training data, can accurately estimate it, which is not the case with non-stationarity. This experiment illustrates the role of the fallback as an explicit
extrapolation assumption. TrustMix does not remove the need for such an
assumption: when the fallback is misspecified, performance under sufficiently
extreme shift is necessarily limited by this choice. Conversely, when relevant
knowledge about extrapolative behaviour is available, it can be incorporated
directly through $p_0$, as demonstrated by the oracle fallback.

\begin{table}[h]
\centering
\small
\caption{Synthetic GP experiments on 10D space, models are given data from $[-1,1]^{10}$ and tested on $[-2,2]^{10}$. Displayed values are  predictive NLL ($\downarrow$), median [Q1, Q3] over 10 seeds. Best method (except for oracle) in bold.}

\begin{tabular}{c|c|c}
\toprule
 & \multicolumn{1}{c}{10D, stationary RBF} & \multicolumn{1}{c}{10D, non-stationary} \\
 & NLL ($\downarrow$) & NLL  ($\downarrow$) \\
\midrule
Base & 18.99 [17.20, 37.15] & 11.29 [5.61, 16.23] \\
$\sigma$-scaling & 12.07 [9.76, 15.65] & 11.94 [9.56, 15.83] \\
Dropout & 2.22 [2.07, 2.59] & 1.34 [1.25, 1.54] \\
$\sigma$-scaling + Dropout & 2.45 [2.22, 2.68] & 1.38 [1.28, 1.57] \\
Dist.-Aware $\sigma$-scaling & 2.18 [2.11, 2.41] & 1.72 [1.52, 1.87] \\
\midrule
TrustMix & 1.44 [1.42, 1.46] & {0.71 [0.69, 0.73]} \\
TrustMix + oracle prior  & \textbf{1.41 [1.41, 1.42]} & \textbf{0.66 [0.65, 0.66]} \\
\midrule
GP oracle & 1.39 [1.39, 1.40] & 0.64 [0.63, 0.64] \\
\bottomrule
\end{tabular}
\label{tab:synthgp_nll_10d}
\end{table}

\begin{table}[h]
\centering
\caption{Median NLL values ($\downarrow$) with first and third quartiles in brackets for different methods on datasets over 10 seeds. Values in bold correspond to best method and ties w.r.t. Wilcoxon test.}
\resizebox{\textwidth}{!}{
\begin{tabular}{c|cc|cccccc}
\toprule
Dataset & \multicolumn{2}{|c}{VPower$_s$} & \multicolumn{2}{|c}{Weather} \\
Metric & NLL ($\downarrow$) & OOD NLL ($\downarrow$) & NLL ($\downarrow$) & OOD NLL ($\downarrow$) \\
\midrule
Base & -0.35 [-0.58, -0.33] & -0.28 [-0.29, -0.20] & -0.09 [-0.15, -0.07] & 1.64 [0.69, 2.50] \\
$\sigma$-scaling & -0.37 [-0.59, -0.34] & -0.28 [-0.30, -0.11] & -0.11 [-0.16, -0.09] & 2.25 [0.92, 3.44] \\
Dropout & 0.08 [0.07, 0.10] & 0.12 [0.09, 0.15] & 0.42 [0.40, 0.49] & \textbf{0.60 [0.52, 0.65]} \\
$\sigma$-scaling + Dropout & -0.28 [-0.30, -0.27] & -0.36 [-0.36, -0.34] & -0.01 [-0.03, 0.02] & \textbf{0.26 [0.24, 0.30]} \\
Dist.-Aware $\sigma$-scaling & -0.37 [-0.59, -0.34] & -0.28 [-0.30, -0.12] & -0.11 [-0.16, -0.09] & 2.25 [0.93, 3.45] \\
\midrule
TrustMix (ours) & \textbf{-0.58 [-0.64, -0.57]} & \textbf{-0.44 [-0.46, -0.35]} & \textbf{-0.29 [-0.30, -0.28]} & \textbf{0.24 [0.10, 0.30]} \\
\bottomrule
\end{tabular}
}
\label{tab:shifts_results}
\end{table}

\begin{table}[h]
\centering

\caption{Median NLL values ($\downarrow$) with first and third quartiles in brackets for different methods on review score prediction over 10 seeds. Bold values are best method and ties w.r.t. Wilcoxon test. Base model here is DistilBERT.}
\footnotesize
\resizebox{\textwidth}{!}{
\begin{tabular}{c|cccc}
\toprule
Dataset & Amazon & Google Play & Trip Advisor & Yelp \\
Metric & NLL ($\downarrow$) & OOD NLL ($\downarrow$) & OOD NLL ($\downarrow$) &  OOD NLL ($\downarrow$) \\
\midrule
Base & 0.66 [0.64, 0.69] & 1.54 [1.50, 1.64] & 1.57 [1.42, 1.70] & 1.72 [1.59, 1.82] \\
$\sigma$-scaling & 0.63 [0.62, 0.65] & 1.42 [1.40, 1.46] & 1.38 [1.31, 1.45] & 1.53 [1.44, 1.56] \\
Dropout & 0.65 [0.64, 0.66] & 1.47 [1.44, 1.56] & 1.54 [1.37, 1.58] & 1.62 [1.50, 1.63] \\
$\sigma$-scaling + Dropout & \textbf{0.64 [0.62, 0.65]} & \textbf{1.41 [1.38, 1.45]} & 1.40 [1.30, 1.43] & \textbf{1.47 [1.41, 1.53]} \\
Dist.-Aware $\sigma$-scaling & \textbf{0.63 [0.62, 0.65]} & \textbf{1.40 [1.38, 1.44]} & 1.36 [1.29, 1.42] & 1.50 [1.41, 1.53] \\
\midrule
TrustMix (ours) & \textbf{0.63 [0.62, 0.65]} & \textbf{1.40 [1.38, 1.44]} & \textbf{1.35 [1.28, 1.41]} & \textbf{1.47 [1.41, 1.50]} \\
\bottomrule
\end{tabular}
}
\label{tab:nlp_results}
\end{table}

\begin{table}[h]
\caption{Median NLL values ($\downarrow$) with first and third quartiles in brackets for different methods on UTK-Face datasets over 10 seeds. Bold values are best method and ties w.r.t. Wilcoxon test. Base model here is ResNet-18.}
\small
\centering
\begin{tabular}{c|cc}
\toprule
Dataset & \multicolumn{2}{|c}{UTK-Face} \\
Metric & NLL ($\downarrow$) & OOD NLL ($\downarrow$) \\
\midrule
Base & 0.39 [0.33, 0.46] & 8.36 [3.94, 12.75] \\
Dropout & 0.38 [0.32, 0.44] &  7.22 [3.60, 10.13] \\
$\sigma$-scaling & 0.36 [0.32, 0.44] & 7.63 [5.26, 12.16] \\
$\sigma$-scaling + Dropout & \textbf{0.34 [0.30, 0.41]} & 7.29 [4.85, 11.12] \\
Dist.-Aware $\sigma$-scaling & \textbf{0.34 [0.31, 0.43]} & 2.12 [1.96, 2.40] \\
\midrule
TrustMix (ours) & \textbf{0.34 [0.30, 0.38]} & \textbf{1.57 [1.43, 1.96]} \\
\bottomrule
\end{tabular}

\label{tab:utk_results}
\end{table}

\subsection{Regression}

We consider tabular (Weather\citep{malinin2021shifts} and VPower$_s$ \citep{malinin2022shifts} datasets), vision (UTK-Face), and NLP tasks. For tabular datasets, we simply train a MLP; for the weather dataset, in-distribution data is drawn from Shifts' canonical training/development split of weather station measurements from a fixed set of times and locations, while out-of-distribution data comes from held-out stations and time periods exhibiting genuine seasonal and geographic shift; for VPower$_s$
  - where the task is predicting the power output of a marine cargo vessel's engine from operating and environmental sensor readings — in-distribution data corresponds to normal, expected vessel operating conditions, while out-of-distribution data corresponds to Shifts' synthetically perturbed operating regimes designed to emulate atypical vessel behaviour.  On UTK-Face, we train a ResNet-18 and introduce distribution shift via ImageNet-C style corruptions \citep{hendrycks2019benchmarking}. For NLP, we train a DistilBERT model to regress review scores from text; scores are normalised to $[0,1]$ and mapped to $\mathbb{R}$ via a logit transform. We treat Amazon reviews \citep{marc_reviews} as in-distribution and evaluate on other domains ( TripAdvisor \citep{tripadvisor_dataset}, Yelp \citep{zhang2015character}, Google Play\footnote{https://huggingface.co/datasets/Gopalatius/google-play-review}) as OOD. Calibration is performed by minimising negative log-likelihood (NLL) on validation data. We report median NLL in Tables~\ref{tab:shifts_results},\ref{tab:nlp_results} and \ref{tab:utk_results} with quartiles over multiple seeds and assess significance using the Wilcoxon test. We bold best method and ties w.r.t. the test. We also present coverage and interval width results in  \ref{sec:coverage}. Across all benchmarks, TrustMix consistently improves OOD NLL while remaining competitive in-distribution, with particularly large gains on UTK-Face. The closest runner-ups are $\sigma$-scaling in combination with MC dropout and the Distance Aware $\sigma$-scaling. However, TrustMix, which relies on the same deviation signal as Distance Aware $\sigma$-scaling, consistently matches or improves upon its NLL. When it comes to coverage results, TrustMix obtains comparable coverage both in and out-of-distribution with other methods on UTK and NLP tasks, while slightly underperforming on Weather and VPower$_s$. All methods (including TrustMix) are calibrated with NLL as the target metric, not coverage or interval width, so these results should be regarded as secondary.

\subsection{Bayesian Optimisation}
\begin{figure}
    \centering
    \includegraphics[width=0.5\linewidth]{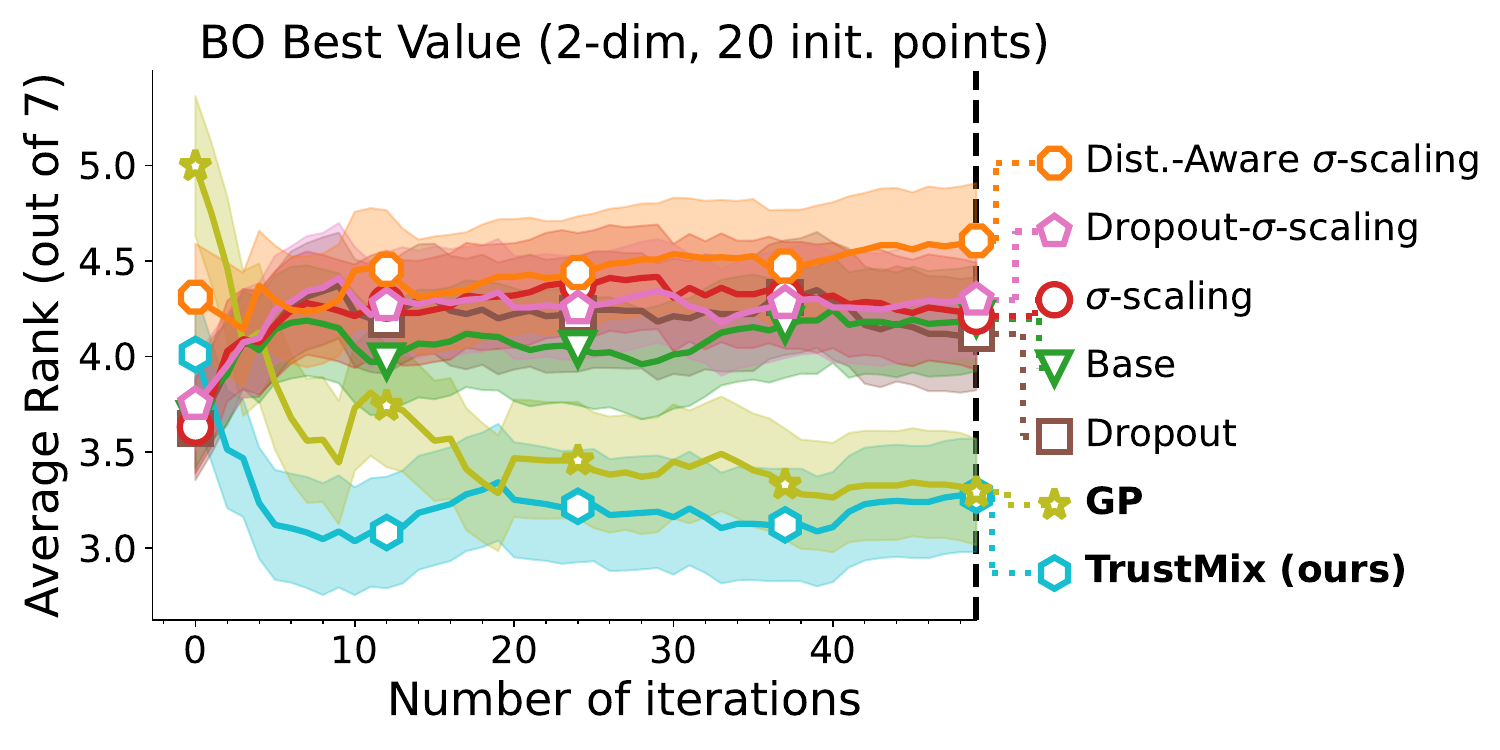}\includegraphics[width=0.5\linewidth]{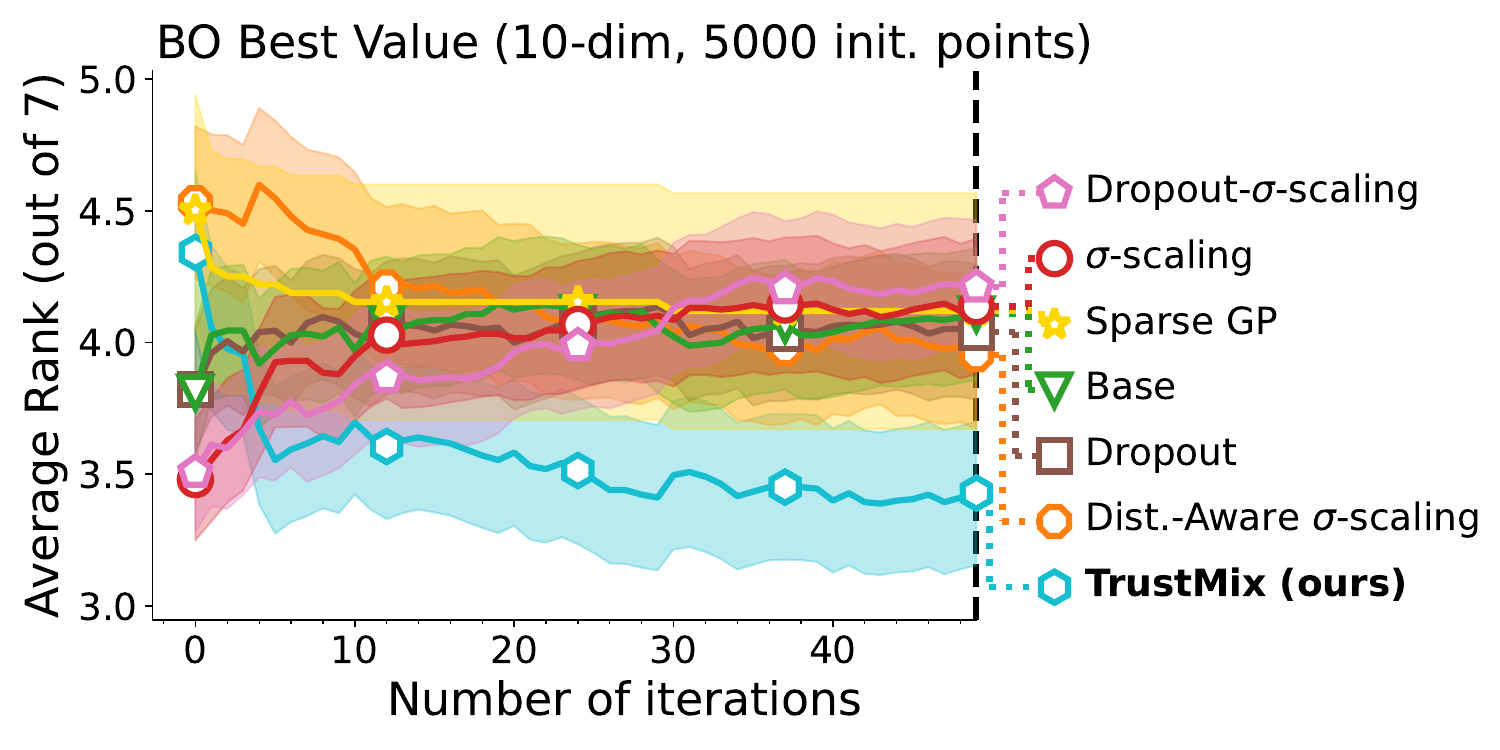}
    \caption{Results on BO experiments. Plots show average rank (lower is better) across all 7 tested functions and all 25 seeds. The plots show how each method’s rank, based on the best value found by each iteration, evolves over time. Best method on last step and ties according to two-sided z-test are in bold.}
    \label{fig:bo}
\end{figure}

To assess TrustMix in downstream decision-making under uncertainty, we compare it with other regression methods when used as a surrogate model for Bayesian optimisation (BO). BO is a natural testbed for uncertainty under distribution shift, as effective optimisation often requires exploring regions far from the observations used to fit the surrogate model. In the problem setting of BO we wish to maximise an unknown black-box function, without access to its derivatives. At each timestep $t$, we can simply query the function at a selected point $x_t$ and observe the function value $f(x_t)$. We then fit a regression model to all datapoints gathered so far $\mathcal{D}_t = \{(x_i, f(x_i))\}_{i=1}^t$ and based on it construct an acquisition function that tells us how promising each point is to query next. In our experiments we use Expected Improvement, defined as:
\begin{equation*}
    a^{\text{EI}}(x|\mathcal{D}_t) = \left(\mu_t(x)-f_t^+\right)
\Phi\left(
\frac{\mu_t(x)-f_t^+}{\sigma_t(x)}
\right)
+
\sigma_t(x)
\phi\left(
\frac{\mu_t(x)-f_t^+}{\sigma_t(x)}
\right),
\end{equation*}
where $f_t^+$ is the best function value so far and $\Phi$ and $\phi$ are Gaussian CDF and PDF respectively and $\mu_t(x)$ and $\sigma_t(x)$ are predicted mean and standard deviation at a given point  $x$ by the fitted regression model. At each timestep we select the point where expected improvement is greatest, that is $x_{t+1} = \arg\max_{x \in \mathcal{X}} a^{\text{EI}}(x|\mathcal{D}_t)$. We test two settings: small-scale, where we start with 20 initial datapoints and operate on a 2-dimensional space and large-scale, where we start with 5000 initial points and operate on a 10-dimensional space. In each setting, we evaluate on seven commonly used benchmark functions: Levy, Michalewicz, Rastrigin, StyblinskiTang, Ackley, Griewank and Rosenbrock. We compare all previously tested regression methods as well as a Gaussian Process regressor in small-scale setting and a Sparse Gaussian Process in large-scale setting. We record the best function value found so far by each method before or on each timestep.
In Figure \ref{fig:bo}, we show the average rank of best values found by each method (lower the better) across all benchmark functions in each setting. We show detailed results in Appendix \ref{ap:bo}.

In a BO problem, there is no clear in-distribution/OOD split, in many cases, we wish to explore points very far from the ones we tested so far, in others, we focus on a promising region close to previous queries. As such, we need a method that exhibits strong OOD performance while being competitive in-distribution. This is reflected in BO results, where TrustMix achieves a lower average rank than the other neural methods early in optimisation and largely maintains this advantage thereafter. In the small-scale setting, TrustMix closely matches the performance  of the gold standard GP model. This model, however, in its standard version is computationally expensive to scale to large datasets in its standard form, hence we rely on SparseGP in the large-scale setting. Note, however, that SparseGP is an approximation to full GP and TrustMix outperforms in terms of average rank across the functions we tested.

\subsection{Classification}
We evaluate TrustMix on three widely used image-classification benchmarks: CIFAR10, CIFAR100 and ImageNet. As the OOD set we use the CIFAR-C and ImageNet-C datasets proposed by \citep{hendrycks2019benchmarking} at the highest corruption severity (corruption level 5). On each dataset, we calibrate two different architectures trained on the corresponding training sets as base models. On CIFAR10/100 we use the canonical WideResNet 28-10 \citep{zagoruyko2016wide} and VGG16BN \citep{simonyan2014very}. On ImageNet we use ResNet152 \citep{he2016deep} and Vision Transformer ViT-B/16 \citep{dosovitskiy2020image}. We report the NLL and Expected Calibration Error (ECE)  in Tables \ref{tab:table_cifar100_posthoc_wrn28_10} and \ref{tab:table_imagenet_posthoc_vit_b_16} as well as in Appendix \ref{ap:class}.

The classification results show that TrustMix is particularly effective at maintaining calibration under distribution shifts, while remaining competitive in-distribution.
Across CIFAR10, CIFAR100, and ImageNet, all methods are calibrated using ID validation data only, making OOD performance a direct test of robustness. Across all datasets and architectures, TrustMix obtains the lowest (or statistically tied with best) OOD ECE, while remaining competitive in-distribution. Improvements in OOD NLL are less consistent: TrustMix is competitive in several settings but is outperformed by other calibration methods on some dataset–architecture combinations. The ImageNet results demonstrate applicability of TrustMix to large datasets and modern architectures.

\begin{table}[h]
\centering
\caption{ECE ($\downarrow$) and NLL ($\downarrow$) on ID and OOD (corruption severity 5) datasets on CIFAR100 with the WideResNet 28-10 architecture. Reported values are means over 3 seeds, with values after $\pm$ being 1.96 times the standard error (half-width of the 95\% confidence interval). Best values and ties w.r.t. two-sided z-test are in bold, second best underlined.}
\footnotesize
\begin{tabular}{c|cc|cc}
\toprule
& \multicolumn{2}{c|}{$10 \times$ECE} & \multicolumn{2}{c}{NLL} \\
Method & ID & OOD & ID & OOD \\
\midrule
Base & 0.63 $\pm$ 0.08 & \underline{2.88 $\pm$ 0.61} & 0.80 $\pm$ 0.01 & \textbf{3.61 $\pm$ 0.36} \\
DAC & 0.39 $\pm$ 0.04 & \underline{2.73 $\pm$ 0.31} & 0.77 $\pm$ 0.01 & \textbf{3.47 $\pm$ 0.13} \\
HTS & \underline{0.39 $\pm$ 0.05} & \underline{2.69 $\pm$ 0.35} & \underline{0.76 $\pm$ 0.00} & \textbf{3.40 $\pm$ 0.15} \\
IOP-IO & \textbf{0.12 $\pm$ 0.02} & \textbf{2.40 $\pm$ 0.35} & \textbf{0.72 $\pm$ 0.01} & \underline{3.81 $\pm$ 0.26} \\
IROVA & 0.47 $\pm$ 0.05 & \textbf{2.50 $\pm$ 0.40} & 1.21 $\pm$ 0.01 & 7.90 $\pm$ 0.64 \\
TS & 0.41 $\pm$ 0.03 & \textbf{2.25 $\pm$ 0.38} & 0.78 $\pm$ 0.00 & \textbf{3.36 $\pm$ 0.18} \\
\midrule
TrustMix (ours) & \underline{0.33 $\pm$ 0.05} & \textbf{1.92 $\pm$ 0.45} & \underline{0.76 $\pm$ 0.01} & \textbf{3.45 $\pm$ 0.21} \\
\bottomrule
\end{tabular}
\label{tab:table_cifar100_posthoc_wrn28_10}
\end{table}

\begin{table}[h]
\centering
\caption{ECE ($\downarrow$) and NLL ($\downarrow$) on ID and OOD (corruption severity 5) datasets on ImageNet with the ViT-B/16 architecture. Reported values are means over 3 seeds, with values after $\pm$ being 1.96 times the standard error (half-width of the 95\% confidence interval). Best values and ties w.r.t. two-sided z-test are in bold, second best underlined.}
\footnotesize
\begin{tabular}{c|cc|cc}
\toprule
& \multicolumn{2}{c|}{$10 \times$ECE} & \multicolumn{2}{c}{NLL} \\
Method & ID & OOD & ID & OOD \\
\midrule
Base & 0.58 $\pm$ 0.02 & \textbf{0.28 $\pm$ 0.02} & 0.84 $\pm$ 0.01 & \textbf{3.25 $\pm$ 0.01} \\
DAC & 0.37 $\pm$ 0.03 & 0.74 $\pm$ 0.07 & \underline{0.79 $\pm$ 0.01} & \textbf{3.24 $\pm$ 0.02} \\
HTS & 0.40 $\pm$ 0.03 & 1.19 $\pm$ 0.04 & \underline{0.79 $\pm$ 0.01} & \underline{3.28 $\pm$ 0.02} \\
IOP-IO-Clip & \textbf{0.14 $\pm$ 0.03} & \underline{0.39 $\pm$ 0.14} & \textbf{0.73 $\pm$ 0.01} & \underline{3.31 $\pm$ 0.03} \\
IROVA & \underline{0.22 $\pm$ 0.04} & 1.13 $\pm$ 0.04 & 1.06 $\pm$ 0.05 & 9.60 $\pm$ 0.06 \\
TS & 0.34 $\pm$ 0.03 & 1.01 $\pm$ 0.06 & \underline{0.79 $\pm$ 0.01} & \textbf{3.27 $\pm$ 0.02} \\
\midrule
TrustMix (ours) & \textbf{0.18 $\pm$ 0.04} & \textbf{0.21 $\pm$ 0.07} & \textbf{0.74 $\pm$ 0.01} & \underline{3.28 $\pm$ 0.03} \\
\bottomrule
\end{tabular}
\label{tab:table_imagenet_posthoc_vit_b_16}
\end{table}

\subsection{Understanding TrustMix} \label{sec:addexp}

\paragraph{Is deviation informative about model reliability?} TrustMix separates the question of what to predict under extreme
extrapolation, encoded by $p_0$, from the question of when the base model
should be trusted, informed by $D(x)$. The latter requires the deviation
signal to be informative about predictive failure. To investigate this
assumption, in Table \ref{tab:table_all_id_ood_trust_mix_no_tp_nll_spearman}, we show Spearman rank correlation between the NLL of the base distribution $p_f(y|x)$ used by TrustMix and the used Mahalanobis deviation signal $D(x)$. We measure this on both in-distribution and out-of-distribution data. We repeat the procedure across all seeds and report the mean correlation, the number of runs with positive correlation, and with statistically significant correlation. While the strength of the association varies across tasks, all measured
rank correlations are positive and statistically significant, supporting the
use of Mahalanobis distance as a deviation signal in the considered settings.

\begin{table}[h]
\centering
\caption{Spearman rank correlation between TrustMix's OOD deviation $D(x)$ and NLL for the base model distribution $p_f(y|x)$ used by TrustMix. Middle column: how many of the evaluations had a positive correlation; right column: how many were significant.}
\footnotesize
\begin{tabular}{l|ccc}
\toprule
Problem & Mean Spearman $\rho$ & \# positive & \# significant ($p < 0.05$) \\
\midrule
CIFAR10 (WideResNet 28-10) & 0.40 & 3/3 & 3/3 \\
CIFAR100 (WideResNet 28-10) & 0.32 & 3/3 & 3/3 \\
ImageNet (ViT-B/16) & 0.39 & 3/3 & 3/3 \\
UTK  & 0.49 & 20/20 & 20/20 \\
Vpower$_s$ & 0.08 & 10/10 & 10/10 \\
Weather & 0.22 & 10/10 & 10/10 \\
NLP -- Amazon + Google Play & 0.22 & 10/10 & 10/10 \\
NLP -- Amazon + Trip Advisor & 0.27 & 10/10 & 10/10 \\
NLP -- Amazon + Yelp & 0.09 & 10/10 & 10/10 \\
\bottomrule
\end{tabular}
\label{tab:table_all_id_ood_trust_mix_no_tp_nll_spearman}
\end{table}

\begin{figure}[h]
    \centering
    \includegraphics[width=\linewidth]{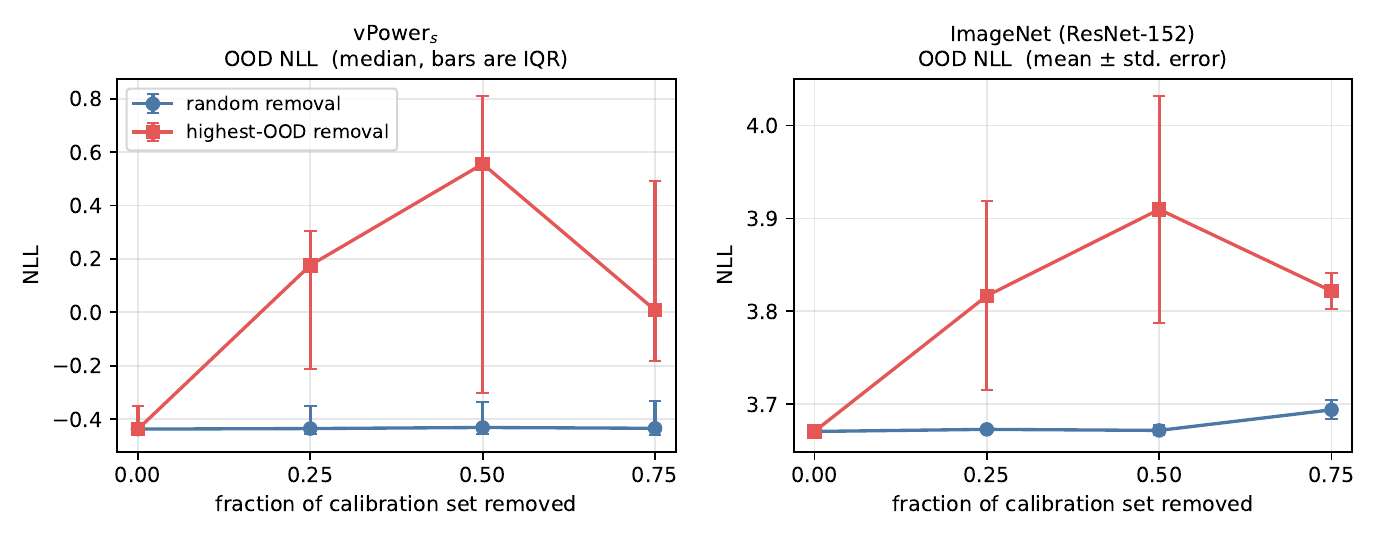}
    \caption{Plots showing the effect of removing validation datapoints on TrustMix. Blue curve shows random removal, where we randomly drop a given fraction of datapoints, whereas the red curve shows the effects of removing a fraction of points with highest OOD signal. }
    \label{fig:valreduce}
\end{figure}

\paragraph{How can trust be learned from ID data alone?}
TrustMix is fitted to the in-distribution validation data only. However, the deviation signal $D(x)$ varies significantly within the in-distribution validation data, allowing TrustMix to learn how trust in the base model should vary with deviation.  We examine the effect of variation within the calibration data in greater detail in Fig. \ref{fig:valreduce}. We reduced the size of the validation sets in two ways: either by randomly eliminating data points or by eliminating the data points with the largest deviations. We evaluated these reductions on two representative problems: regression on vPower$_s$ and classification on ImageNet. Eliminating validation data with large deviations typically produces substantially greater degradation in OOD performance compared to eliminating the same proportion of validation data at random. These results suggest that TrustMix utilizes the variability of deviation among in-distribution calibration examples (particularly those with large deviations) to determine the rate at which trust in the base model should be reduced.

\begin{figure}[h]
    \centering
    \includegraphics[width=\linewidth]{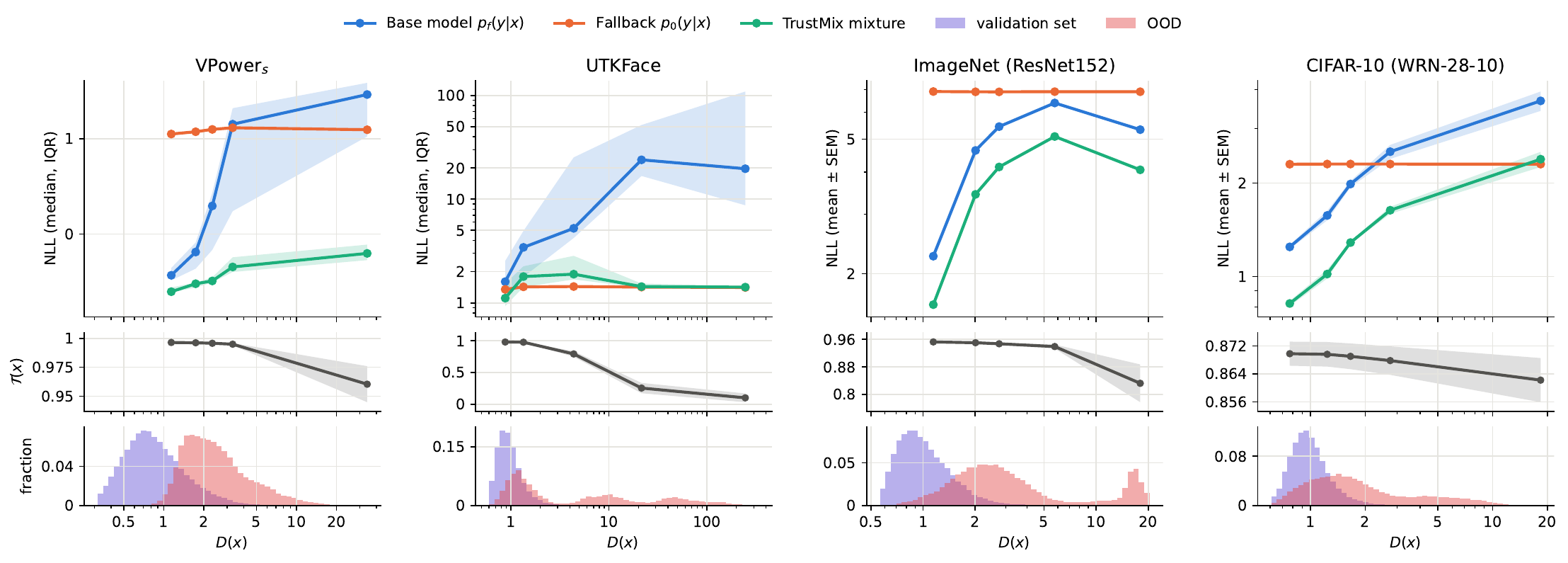}
    \caption{Behaviour of TrustMix on various OOD dataset. Top row: predictive NLL of the base model ($p_f(y|x)$), fallback (\(p_0(y|x)\)), and resulting TrustMix mixture $p(y|x)$. Middle row: learned trust \(\mathcal{T}(x)\). Bottom row: distribution of \(D(x)\) over validation and OOD evaluation datasets. Lines and shaded areas in middle plot correspond to mean and standard errors over seeds. Top and Middle plots are made by binning the deviation signal on OOD datasets into 5 bins, plots show respective summary values over each bin.}
    \label{fig:trust_curves}
\end{figure}

\paragraph{How good is the default fallback?} In Figure \ref{fig:trust_curves} we show how NLL changes of TrustMix's predictive distribution changes on various OOD datasets as the input deviates from training distribution. We bin the datapoints from OOD datasets into 5 bins depending on their deviation signal and display summary values for each bin. Alongside, we plot the NLL of the base distribution $p_f(y|x)$ and that of the fallback $p_0(y|x)$. We also show the average trust gate value $\mathcal{T}(x)$ assigned at each deviation level in the  middle plots and distribution of OOD and validation datapoints in the bottom plots.

We see that the base model $p_f(y|x)$ gets worse on large deviation examples, while fallback is roughly flat. As such, on multiple datasets, the the base model eventually becomes worse than the fallback. This shows that reverting to a simple, input-independent fallback can meaningfully improve on base model in far OOD scenarios. In particular on the UTK dataset, fallback is substantially better than base model and TrustMix leverages that by quickly collapsing to it. However, on ImageNet, the NLL of base model gets closer to the one of fallback, but is never worse. Even then, by mixing the two distribution, TrustMix achieves better NLL than either of them. This reveals one important property of the fallback: on its own it does not necessarily needs to be better than the base model to allow TrustMix to make meaningful improvements via mixing.

\paragraph{TrustMix with a different deviation signal $D(x)$} Throughout the paper, we use Mahalanobis distance as our default deviation signal. However, TrustMix can in principle be used with alternative signals, such as logit energy \citep{liu2020energy}. We conducted an experiment on  ImageNet, where we replaced the deviation signal in TrustMix and used Logit Energy, normalised by smallest training set energy so that the signal lies in $[0, + \infty]$. We compare the Mahalanobis distance-based deviation signal defined in Appendix \ref{app:mah} with an energy-based deviation signal proposed by \citep{liu2020energy}:
$
    E(z(x)) = - \log \sum_{k=1}^C e^{z_k(x)},
$
where $z_h(x)$ are the logits for the $k$th class for a given input $x \in \mathcal{X}$ as given by the base model and $C$ is the number of classes.  As $E(z)$ can be both negative and positive, to convert it into a non-negative deviation signal that can be used by TrustMix we transform it as:

\begin{equation*}
    D^{\textrm{energy}}(x) = \frac{E(x) - E^{\textrm{min}}}{\hat{E} - E^{\textrm{min}}},
\end{equation*}
with $E^{\textrm{min}} = \min_{x \in \mathcal{D}_{\textrm{train}}} E(x)$ and $\hat{E} = \frac{1}{|\mathcal{D}_{\textrm{train}}|}\sum_{x \in \mathcal{D}_{\textrm{train}}} E(x) $. We instantiate TrustMix with $D^{\textrm{energy}}(x)$ instead of Mahalanobis and show the result in Table \ref{tab:table_imagenet_tm_energy_vit_b_16}. We see that this makes little change with ViT-B/16 as a base model, yet meaningfully impacts OOD performance with ResNet 152 used as base. This is consistent with earlier literature reporting Mahalanobis to be generally stronger OOD detector than Energy \citep{mukhoti2023deep}.

\paragraph{TrustMix operating in probability space}
Earlier we explained that operating in logits space should give TrustMix more flexibility than operating in pure probability space. We now verify this claim experimentally; in Table \ref{tab:table_imagenet_tm_fallback_vs_probmixing_vit_b_16} we show a comparison of performance on ImageNet of standard TrustMix with a TrustMix operating in probability space, where $p_f$ denotes the class probabilities produced by the base model (with learnable temperature) and $p_0$ is a uniform categorical distribution over classes. Note that when $\mathcal{T}(x)=1$, both models have the same flexibility. As such, the only difference is how the mixing is conducted. We see that, as predicted, mixing in probability space achieves much worse in-distribution performance. out-of-distribution, it achieves slightly better NLL on ResNet 152 at the cost of substantially worse ECE.

\begin{table}[h]
\centering
\caption{ECE ($\downarrow$) and NLL ($\downarrow$) on ID and OOD (corruption severity 5) datasets on ImageNet with the ViT-B/16 and ResNet152 architectures, comparing TrustMix's default Mahalanobis OOD deviation signal against an Energy signal. Reported values are means over 3 seeds, with values after $\pm$ being 1.96 times the standard error (half-width of the 95\% confidence interval). Best values and ties w.r.t. two-sided z-test are in bold.}
\footnotesize
\begin{tabular}{c|cc|cc}
& \multicolumn{4}{c}{ViT-B/16} \\
\toprule
& \multicolumn{2}{c|}{$10 \times$ECE} & \multicolumn{2}{c}{NLL} \\
Method & ID & OOD & ID & OOD \\
\midrule
TrustMix (ours) & \textbf{0.18 $\pm$ 0.04} & \textbf{0.21 $\pm$ 0.07} & \textbf{0.74 $\pm$ 0.01} & \textbf{3.28 $\pm$ 0.03} \\
TrustMix (Energy) & \textbf{0.18 $\pm$ 0.04} & \textbf{0.23 $\pm$ 0.07} & \textbf{0.74 $\pm$ 0.01} & \textbf{3.29 $\pm$ 0.03} \\
\bottomrule
& \multicolumn{4}{c}{ResNet 152} \\
\toprule
& \multicolumn{2}{c|}{$10 \times$ECE} & \multicolumn{2}{c}{NLL} \\
Method & ID & OOD & ID & OOD \\
\midrule
TrustMix (ours) & \textbf{0.17 $\pm$ 0.03} & \textbf{0.69 $\pm$ 0.23} & \textbf{0.68 $\pm$ 0.01} & \textbf{3.67 $\pm$ 0.01} \\
TrustMix (Energy) & \textbf{0.17 $\pm$ 0.03} & \underline{1.03 $\pm$ 0.02} & \textbf{0.68 $\pm$ 0.01} & \underline{3.72 $\pm$ 0.03} \\
\bottomrule
\end{tabular}
\label{tab:table_imagenet_tm_energy_vit_b_16}
\end{table}

\begin{table}[h]
\centering
\caption{ECE ($\downarrow$) and NLL ($\downarrow$) on ID and OOD (corruption severity 5) datasets on ImageNet with the ViT-B/16 and ResNet152 architectures, comparing TrustMix's logit-space fallback (default) against probability-space mixing. Reported values are means over 3 seeds, with values after $\pm$ being 1.96 times the standard error (half-width of the 95\% confidence interval). Best values and ties w.r.t. two-sided z-test are in bold.}
\footnotesize
\begin{tabular}{c|cc|cc}
& \multicolumn{4}{c}{ViT-B/16} \\
\toprule
& \multicolumn{2}{c|}{$10 \times$ECE} & \multicolumn{2}{c}{NLL} \\
Method & ID & OOD & ID & OOD \\
\midrule
TrustMix (Logit Space) & \textbf{0.18 $\pm$ 0.04} & \textbf{0.21 $\pm$ 0.07} & \textbf{0.74 $\pm$ 0.01} & \textbf{3.28 $\pm$ 0.03} \\
TrustMix (Prob. Space) & \underline{0.35 $\pm$ 0.02} & \underline{1.04 $\pm$ 0.06} & \underline{0.79 $\pm$ 0.01} & \textbf{3.27 $\pm$ 0.02} \\
\bottomrule
& \multicolumn{4}{c}{ResNet 152} \\
\toprule
& \multicolumn{2}{c|}{$10 \times$ECE} & \multicolumn{2}{c}{NLL} \\
Method & ID & OOD & ID & OOD \\
\midrule
TrustMix (Logit Space) & \textbf{0.17 $\pm$ 0.03} & \textbf{0.69 $\pm$ 0.23} & \textbf{0.68 $\pm$ 0.01} & \underline{3.67 $\pm$ 0.01} \\
TrustMix (Prob. Space) & \underline{0.34 $\pm$ 0.02} & \underline{1.84 $\pm$ 0.01} & \underline{0.74 $\pm$ 0.01} & \textbf{3.62 $\pm$ 0.02} \\
\bottomrule
\end{tabular}
\label{tab:table_imagenet_tm_fallback_vs_probmixing_vit_b_16}
\end{table}

\section{Conclusions}

In this work, we explicitly separated two questions that are often conflated: when should a model be trusted, and what should we predict when trust in the model is low? TrustMix addresses the first question by learning a monotonic mapping from a deviation signal to model trust. It addresses the second through a user-specified fallback distribution, which defines the predictive behaviour as trust in the base model vanishes. Importantly, TrustMix does not make a binary choice between trusting the base model and deferring to the fallback. Instead, it smoothly interpolates between their predictive distributions, allowing predictions to rely partially on the base model and partially on the fallback.

Across regression benchmarks, TrustMix achieved strong OOD predictive likelihood while remaining competitive in-distribution. In classification, its most consistent advantage was OOD calibration, achieving the lowest or statistically tied ECE across the evaluated settings, although improvements in OOD NLL were task-dependent. TrustMix also performed competitively as a neural surrogate for Bayesian optimisation.

The main limitation of TrustMix is its dependence on the choice of fallback distribution. If the fallback is poorly specified for a given task, collapsing towards it under large distribution shifts can degrade predictive performance. TrustMix therefore does not eliminate assumptions about extrapolation; rather, it makes them explicit and controllable. A promising direction for future work is to design task-specific fallback distributions for applications in which extrapolation beyond the training domain is particularly important.


\vskip 0.2in
\bibliographystyle{tmlr}
\bibliography{sample}
\appendix
\newpage

\section{Proof of Proposition \ref{prop:monotone}}
\label{ap:monotone_proof}

\textit{\textbf{Proposition \ref{prop:monotone} (Architectural monotonicity) }.
Let $g(D) = W_L\,\phi(W_{L-1}\,\phi(\cdots\phi(W_1 D + b_1)\cdots) +
b_{L-1}) + b_L$, where every $W_l$ has positive entries and $\phi$ is
non-decreasing and unbounded. Then $g(\cdot)$ is non-decreasing in $D$, and
hence $\text{sigmoid}(-g(D(x);\psi))$ is
non-increasing in $D(x)$ and as
$D(x)\to\infty$, we get $\mathcal{T}(x)\to 0$.}
\\
\begin{proof}
We prove the claim by induction over the layers. Define
$$
g_1(D)=W_1D+b_1,
$$
and, for $l=2,\ldots,L$,
$$
g_l(D)=W_l\phi(g_{l-1}(D))+b_l,
$$
Let $D_1\leq D_2$. Since every entry of $W_1$ is positive, we have that coordinatewise:

$$
W_1D_1+b_1\leq W_1D_2+b_1,
$$
so
$$
g_1(D_1)\leq g_1(D_2).
$$
Thus $g_1$ is coordinatewise non-decreasing in $D$. Now suppose, inductively, that $g_{l-1}$ is coordinatewise non-decreasing. Then for
$D_1\leq D_2$,

$$
g_{l-1}(D_1)\leq g_{l-1}(D_2).
$$
Because $\phi$ is non-decreasing,
$$
\phi(g_{l-1}(D_1))
\leq
\phi(g_{l-1}(D_2)).
$$
Multiplying by $W_l$, whose entries are positive, preserves the coordinatewise inequality, and hence

$$
g_l(D_1)
=
W_l\phi(g_{l-1}(D_1))+b_l
\leq
W_l\phi(g_{l-1}(D_2))+b_l
=
g_l(D_2).
$$Therefore $g_l$ is also coordinatewise non-decreasing. By induction, $g=g_L$ is non-decreasing in $D$. It remains to prove the limit. Since every entry of $W_1$ is positive,

$$
g_1(D)=W_1D+b_1\longrightarrow +\infty
\qquad\text{coordinatewise as }D\to\infty.
$$ Because $\phi$ is non-decreasing and unbounded, this implies

$$
\phi(g_1(D))\longrightarrow+\infty
\qquad\text{coordinatewise}.
$$ Since every entry of $W_2$ is positive,

$$
g_2(D)=W_2\phi(g_1(D))+b_2\longrightarrow+\infty.
$$ Repeating this argument layer by layer gives

$$
g(D)=g_L(D)\longrightarrow+\infty
\qquad\text{coordinatewise as }D\to\infty.
$$ Finally, $\mathrm{sigmoid}(x)=\frac{1}{1 + e^{-x}}$ is increasing and
$\mathrm{sigmoid}(x)\to 0 $ as $x\to -\infty$. Hence

$$
\mathcal{T}(D) = \mathrm{sigmoid}(-g(D))\longrightarrow 0
$$ as $D \to \infty$. Moreover since sigmoid is increasing and $g(D)$ is non-decreasing, we have that $\mathcal{T}(D)$ is non-increasing in $D$.
\end{proof}

\newpage
\section{Experimental Details} \label{ap:expdetails}

All experiments were executed on compute nodes equipped with NVIDIA A40 GPUs, each providing 48\,GB of GPU memory.

We adopt a unified experimental protocol across methods whenever feasible in order to isolate methodological differences. Stochastic Dropout inference relies on five Monte Carlo forward passes, with the dropout rate fixed to $0.3$ throughout. Aside from the final prediction layer—which is method-dependent — all approaches share an identical backbone architecture. We use Adam for optimisation unless stated otherwise.

For regression tasks, we first train the model's mean head by MSE. After this training is completed, we freeze all the layers except for the weights of uncertainty  head and train it to maximise Gaussian likelihood. The uncertainty head predicts the log of standard deviation. When combining with dropout, we take the output mean to be average mean head output, and variance is the sum of mean's head variance and  average variance given by uncertainty head. For classification tasks, we simply train the model to output logits for each class with standard cross entropy loss.

When calibrating all models with all methods, we first precompute base model outputs and then calibrate them. This allows us to save time and memory and thus when calibrating we always use full batch gradient descent. When calibrating, we optimise log likelihood until the change over 10 steps is less than 0.0001.

For all datasets and methods, training is governed by a patience-based early-stopping criterion: if the final epoch yields the lowest validation loss observed so far, training is continued until no additional improvement is detected.

\subsection{Regression -- Weather and VPower$_s$}

For regression on Weather and VPower$_s$ datasets, we use a fully connected MLP backbone with two hidden layers of 256 units. Model is trained with  a learning rate of $10^{-3}$ for the mean head and  variance head. Mean head is trained for one full epoch with batch size 512 and variance head is trained for 10 epochs with full batch of precomputed embeddings. Validation MSE for main model and NLL for variance head are computed every epoch, and the model parameters corresponding to the lowest validation loss are selected for evaluation. Training takes up to 10 minutes.

\subsection{Regression -- UTK}
For regression on UTK-Face, we use a ResNet-18 backbone \citep{he2016deep} initialised from random weights, followed by a single linear output layer. Model is trained for 50 epochs with a batch size of 64 using Adam at a learning rate of $10^{-3}$ for both mean and uncertainty heads. We use data augmentation introduced by \citep{zagoruyko2016wide}. Validation MSE for main model and NLL for variance head is computed every epoch, and the model parameters corresponding to the lowest validation loss are selected for evaluation. Training takes up to 20 minutes.

\subsection{Regression -- NLP tasks}
For regression on NLP tasks, we use DistilBERT \citep{sanh2019DistilBERT}, where we add a linear layer on top. The review scores for each dataset are first normalised to $[0,1]$ range and then transformed via a logit transform. Before transformation, the scores are clipped to $[0.01, 0.99]$ to avoid infinite scores.  Model is trained for 50 epochs with a batch size of 128 using Adam at a learning rate of $10^{-5}$ for both mean and uncertainty heads. Training can take up to 4 hours.

\subsection{Classification -- CIFAR-10/100}
For classification experiments on CIFAR-10 and CIFAR-100 we train all base models from scratch. We use data augmentation following the standard procedure introduced by \citep{zagoruyko2016wide}. For WideResNet 28-10, optimisation is performed using stochastic gradient descent with momentum $0.9$ and weight decay $5 \times 10^{-4}$ over 200 epochs. The learning rate is initialised to $0.1$ and reduced by a factor of $5$ after epochs 60, 120, and 180. For VGG16BN, we train with momentum $0.9$  and weight decay $5 \times 10^{-4}$ for 100 epochs, initialise the learning rate to $0.03$ and divide by 5 at epochs 60 and 80. All experiments use a batch size of 128. Training can take up to 3 hours.

\subsection{Classification -- ImageNet}
On ImageNet, we use pretrained models provided by torchvision \citep{paszke2019pytorch} that were trained  on the ImageNet trainset. We then take the original ImageNet validation set consisting of 50{,}000 examples and split it randomly (seeded with a given random seed) in half to create a calibration set and evaluation set. For OOD, for each evaluation image, we randomly select one type of corruption proposed by \citep{hendrycks2019benchmarking} and apply it at the highest severity.

\subsection{Bayesian Optimisation}
In Bayesian Optimisation, we need to remain particularly sample efficient, so we adopt a slightly different approach compared to supervised learning. We keep two models; one trained on a fraction of available data with the rest set aside for validation and the other trained on full dataset. As validation we randomly select 25\% of points excluding the initial design, together with 5 points from random design in small-scale setting and 20 points in large-scale setting. This is because the initial design is likely to have a completely different distribution from that of points gathered during BO. By restricting the number of data points from the initial design entering the validation we effectively focus more on the BO region. Before fitting the model, we apply the Yeo-Johnson transform for normality to the observed values. We record the average loss achieved by the first model on the training set when the validation set loss is best. Then when training the second model, without the validation set, we simply stop whenever we achieve an average loss that is better or equal. For $\sigma$-scaling or TrustMix, all calibration parameters are selected using the first model, using the validation set for calibration and then simply copied over to the full model. We train both models with a learning rate of $10^{-3}$ for 1500 epochs at first and then we reuse them at further iterations while only training them for a further 100 epochs. We do a full model reset every 10 iterations. One run takes up to 12 minutes.

\newpage
\section{Details of Mahalanobis Computations}
\label{app:mah}
Given a pretrained feature extractor $\Phi(x)$, Mahalanobis distance quantifies the representation space deviation as:
\begin{equation*}
    D(x) = (\Phi(x) - \mu_{\mathrm{train}})^\top \Sigma_{\mathrm{train}}^{-1} (\Phi(x) - \mu_{\mathrm{train}}),
\end{equation*}
where
{
\begin{align*}
\mu_{\mathrm{train}} &= \frac{1}{n} \sum_{x \in \mathcal{D}_{\mathrm{train}}} \Phi(x),  \qquad
\Sigma_{\mathrm{train}} = \frac{1}{n} \sum_{x \in \mathcal{D}_{\mathrm{train}}} (\Phi(x) - \mu_{\mathrm{train}})(\Phi(x) - \mu_{\mathrm{train}})^\top
\end{align*}}denote the empirical mean and covariance of the training embeddings with $n = |\mathcal{D}_{\mathrm{train}}|$.
Both quantities can be computed efficiently using a single forward pass over the training set. Computing Mahalanobis distance requires inverting the covariance $\Sigma_{\mathrm{train}}$, which is performed once per trained model and incurs negligible overhead relative to training. Importantly, $\Sigma_{\mathrm{train}}$ is defined over the feature dimension rather than the dataset size, so its size does not scale with the number of training examples. In practice, covariance inversion can be efficiently handled using standard techniques such as Lanczos method \citep{lanczos1950iteration}. Also, in practice, we find that there is no need to compute $\mu_{\mathrm{train}}$ and $\Sigma_{\mathrm{train}} $ over the entire training set. For larger training sets, we find it sufficient to compute this only on small fraction of training set. Throughout our experiments we use 1\% of all training data on ImageNet and the Amazon data in NLP task. For remaining tasks, we use the entire training set.

In deeper architectures, we find that computing Mahalanobis distance with intermediate features instead of only the last layer features, improves the quality of the OOD signal. Assuming the architectures consists of $m$ blocks after which we extract intermediate representation and the full feature extractor can be rewritten as $\Phi(x) = \Phi^m(\Phi^{m-1}(\dots \Phi^{2}(\Phi^1(x)) \dots))$, we define the per-block Mahalanobis distance as:
\begin{equation*}
    D^j(x) = (\Phi^j(x) - \mu^j_{\mathrm{train}})^\top (\Sigma^j_{\mathrm{train}})^{-1} (\Phi^j(x) - \mu^j_{\mathrm{train}}),
\end{equation*}
where analogously:
\begin{align*}
\mu^j_{\mathrm{train}} &= \frac{1}{n} \sum_{x \in \mathcal{D}_{\mathrm{train}}} \Phi^j(x),  \qquad
\Sigma^j_{\mathrm{train}} = \frac{1}{n} \sum_{x \in \mathcal{D}_{\mathrm{train}}} (\Phi^j(x) - \mu^j_{\mathrm{train}})(\Phi^j(x) - \mu^j_{\mathrm{train}})^\top.
\end{align*}

The final deviation signal is the average per-block distance:
\begin{equation*}
    D(x) = \frac{1}{m} \sum_{j=1}^m  D^j(x) .
\end{equation*}
In Table \ref{tab:mah_layers} below we detail which intermediate layers we use for Mahalanobis distance computations for each architecture among the experiments. Whenever we compare with DAC in classification, we use the same intermediate layer for Mahalanobis computation as for density estimation in DAC for fair comparison.
\begin{table}[H]
\centering
\caption{Intermediate representations used for Mahalanobis distance computations across architectures and experimental settings. For convolutional networks, we extract features after major stages or blocks of the network. For transformer-based models, we use the hidden representations after each transformer block together with the final normalization layer.}
\label{tab:mah_layers}
\begin{tabular}{lll}
\toprule
\textbf{Experiment} & \textbf{Architecture} & \textbf{Feature representations used} \\
\midrule

CIFAR10/100 &
WRN-28-10 &
Outputs after the three residual groups \\

&
ResNet-110 &
Outputs after the three residual groups \\

 &
VGG16BN &
Outputs after convolutional blocks 2, 3, 4, and 5, \\
&& together with the final convolutional feature map \\

ImageNet &
ResNet-152 &
Outputs after residual stages 1--4 \\

 &
ViT-B/16 &
Representations after all 12 transformer encoder blocks \\
& & and the final layer normalization \\

&
DenseNet-169 &
Outputs after transition layers 1--3 and the final \\
&& pooled representation before classification \\

UTK &
ResNet-18 &
Outputs after residual stages 1--4 \\

NLP &
DistilBERT &
Final embeddings only \\

UCI & MLP & Final embeddings only \\

\bottomrule
\end{tabular}
\end{table}

\newpage

\section{Details on TrustMix in classification} \label{ap:tm_class}
In classification with $C$ classes, we treat the logits space as the output space of base model as $\mathcal{Y} = \mathbb{R}^C$. As such, the fallback parameters are now vectors, that is $\mu_0 \in \mathbb{R}^C$ and $\sigma^2_0 \in \mathbb{R}^C$ and the fallback distribution is a diagonal Gaussian over logits space $p_0(y) = \mathcal{N}(y; \mu_0; \textrm{diag}(\sigma_0^2))$. We also apply an affine transformation to the logits first, that is we take $\alpha f(x) + \beta$ as base logits.
$\mu_0$ and $\sigma^2_0$ are then set relatively to base model, by the elementwise vector equations below:
\begin{align*}
\begin{split}
\label{eq:fallback_params1}
    \mu_0 &= \frac{1}{|D_{\textrm{train}}|} \sum_{(x,y) \in D_{\textrm{train}}} f(x)  \\
    \sigma^2_0 &= \frac{1}{|D_{\textrm{train}}|} \sum_{(x,y) \in D_{\textrm{train}}} (f(x)  - \mu_0)^2.
\end{split}
\end{align*}
Denoting $\theta=(\psi, \rho, \alpha, \beta)$, the predictive distribution over logits is also defined by element-wise vector equations:
\begin{align*}
\mu(x;\theta) &= \mathcal{T}_\psi(x) (\alpha f(x) + \beta) + (1-\mathcal{T}_\psi(x)) \mu_0, \\
\sigma^2(x; \theta) &= \mathcal{T}_\psi(x)\sigma^2_f + (1-\mathcal{T}_\psi(x))\sigma_0^2 + \mathcal{T}_\psi(x)(1-\mathcal{T}_\psi(x))(\alpha f(x) + \beta - \mu_0)^2,
\end{align*}
where $\mu(x;\theta) \in \mathbb{R}^C$ and $\sigma^2(x; \theta,) \in \mathbb{R}^C$, but $\mathcal{T}_\psi(x) \in \mathbb{R}$ is a scalar. Since the base model outputs logits but not variance over logits, we just set $\sigma_f = \rho$. Finally, the equation for predictive distribution over classes:
\begin{equation*}
    p(y \mid x;\theta) = \textrm{softmax}\left(\frac{\mu(x;\theta)}{\sqrt{1 + \frac{\pi^2}{8}\sigma^2(x;\theta)}} \right),
\end{equation*}
means we apply softmax to the vector of logits. In other words, if we let $[v]_i$ denote the $i$th entry of vector $v$, then the $i$th element of logits vector $l \in \mathbb{R}^C$ is:
\begin{equation*}
    [l]_i = \frac{[\mu(x;\theta)]_i}{\sqrt{1 + \frac{\pi^2}{8}[\sigma^2(x;\theta)]_i}}
\end{equation*}
and the $i$th entry after softmax is applied is:
\begin{equation*}
    \left[\textrm{softmax}\left(\frac{\mu(x;\theta)}{\sqrt{1 + \frac{\pi^2}{8}\sigma^2(x;\theta)}} \right)\right]_i = \frac{e^{[l]_i}}{\sum_{j =1,\dots,C} e^{[l]_j}}
\end{equation*}
\newpage

\section{Coverage and Interval Width Results} \label{sec:coverage}

\begin{table}[h]
\centering
\caption{95\% predictive interval coverage and width for all methods on VPower$_s$ and Weather datasets. Displayed values are averages over 10 seeds with standard errors after $\pm$ sign. Best values and ties according to z-test in bold, second best are underlined.}
\begin{tabular}{c|cc|cc}
\toprule
Dataset & \multicolumn{2}{|c}{VPower$_s$} & \multicolumn{2}{|c}{Weather} \\
Metric & Coverage  & OOD Coverage & Coverage & OOD Coverage\\
\midrule
Base & \underline{0.96 $\pm$ 0.01} & \textbf{0.93 $\pm$ 0.02} & 0.97 $\pm$ 0.00 & \underline{0.89 $\pm$ 0.02} \\
$\sigma$-scaling & \textbf{0.95 $\pm$ 0.00} & \textbf{0.91 $\pm$ 0.02} & \underline{0.95 $\pm$ 0.00} & 0.87 $\pm$ 0.02 \\
Dropout & 0.99 $\pm$ 0.00 & \underline{1.00 $\pm$ 0.00} & 1.00 $\pm$ 0.00 & \underline{0.99 $\pm$ 0.00} \\
$\sigma$-scaling + Dropout & \textbf{0.95 $\pm$ 0.00} & \textbf{0.97 $\pm$ 0.00} & \textbf{0.95 $\pm$ 0.00} & \textbf{0.93 $\pm$ 0.01} \\
Dist.-Aware $\sigma$-scaling & \textbf{0.95 $\pm$ 0.00} & \textbf{0.91 $\pm$ 0.02} & \underline{0.95 $\pm$ 0.00} & 0.87 $\pm$ 0.02 \\
\midrule
TrustMix (ours) & \textbf{0.94 $\pm$ 0.00} & \underline{0.91 $\pm$ 0.02} & 0.94 $\pm$ 0.00 & 0.85 $\pm$ 0.01 \\
\bottomrule
\toprule
Metric & Width ($\downarrow$) & OOD Width ($\downarrow$) & Width ($\downarrow$) & OOD Width ($\downarrow$) \\
\midrule
Base & \underline{0.75 $\pm$ 0.10} & \underline{0.67 $\pm$ 0.09} & 1.04 $\pm$ 0.05 & 1.12 $\pm$ 0.10 \\
$\sigma$-scaling & \underline{0.67 $\pm$ 0.07} & \textbf{0.60 $\pm$ 0.06} & \underline{0.91 $\pm$ 0.03} & \underline{0.98 $\pm$ 0.08} \\
Dropout & 1.69 $\pm$ 0.03 & 1.75 $\pm$ 0.05 & 2.41 $\pm$ 0.13 & 2.77 $\pm$ 0.33 \\
$\sigma$-scaling + Dropout & 0.80 $\pm$ 0.02 & 0.83 $\pm$ 0.02 & 1.05 $\pm$ 0.04 & 1.21 $\pm$ 0.12 \\
Dist.-Aware $\sigma$-scaling & \underline{0.67 $\pm$ 0.07} & \textbf{0.60 $\pm$ 0.06} & \underline{0.91 $\pm$ 0.03} & \underline{0.98 $\pm$ 0.08} \\
\midrule
TrustMix (ours) & \textbf{0.53 $\pm$ 0.01} & \textbf{0.54 $\pm$ 0.04} & \textbf{0.72 $\pm$ 0.00} & \textbf{0.71 $\pm$ 0.04} \\
\bottomrule
\end{tabular}
\label{tab:results1}
\end{table}

\begin{table}[h]
\centering
\caption{95\% predictive interval coverage and width for all methods on NLP tasks. Displayed values are averages over 10 seeds with standard errors after $\pm$ sign. Best values and ties according to z-test in bold, second best are underlined.}
\begin{tabular}{c|cccc}
\toprule
Dataset & Amazon & Google Play & Trip Advisor & Yelp\\
Metric & Coverage  & OOD Coverage  &  OOD Coverage  &  OOD Coverage  \\
\midrule
Base & 0.93 $\pm$ 0.00 & \underline{0.86 $\pm$ 0.02} & 0.77 $\pm$ 0.03 & \underline{0.79 $\pm$ 0.02} \\
$\sigma$-scaling & \textbf{0.95 $\pm$ 0.00} & \textbf{0.90 $\pm$ 0.01} & \textbf{0.85 $\pm$ 0.02} & \textbf{0.85 $\pm$ 0.01} \\

Dropout & \underline{0.94 $\pm$ 0.00} & \underline{0.88 $\pm$ 0.01} & 0.80 $\pm$ 0.03 & \underline{0.82 $\pm$ 0.02} \\
$\sigma$-scaling + Dropout & \textbf{0.95 $\pm$ 0.00} & \textbf{0.90 $\pm$ 0.01} & \underline{0.84 $\pm$ 0.02} & \textbf{0.85 $\pm$ 0.01} \\
Dist.-Aware $\sigma$-scaling & \textbf{0.95 $\pm$ 0.00} & \textbf{0.91 $\pm$ 0.01} & \textbf{0.86 $\pm$ 0.01} & \textbf{0.86 $\pm$ 0.01} \\
\midrule
TrustMix (ours) & \textbf{0.95 $\pm$ 0.00} & \textbf{0.91 $\pm$ 0.01} & \textbf{0.86 $\pm$ 0.01} & \textbf{0.87 $\pm$ 0.01} \\
\bottomrule
\toprule
Dataset & Amazon & Google Play & Trip Advisor & Yelp\\
Metric & Width  & OOD Width  &  OOD Width  &  OOD Width  \\
\midrule
Base & \textbf{1.81 $\pm$ 0.03} & \textbf{2.39 $\pm$ 0.06} & \textbf{1.93 $\pm$ 0.06} & \textbf{2.11 $\pm$ 0.06} \\
$\sigma$-scaling & 2.04 $\pm$ 0.02 & 2.69 $\pm$ 0.04 & \underline{2.17 $\pm$ 0.04} & 2.37 $\pm$ 0.04 \\
Dropout & \underline{1.90 $\pm$ 0.03} & \underline{2.48 $\pm$ 0.06} & \textbf{2.01 $\pm$ 0.06} & \underline{2.20 $\pm$ 0.06} \\
$\sigma$-scaling + Dropout & 2.03 $\pm$ 0.02 & 2.66 $\pm$ 0.03 & \underline{2.15 $\pm$ 0.04} & 2.36 $\pm$ 0.04 \\
Dist.-Aware $\sigma$-scaling & 2.07 $\pm$ 0.02 & 2.74 $\pm$ 0.04 & \underline{2.21 $\pm$ 0.04} & 2.41 $\pm$ 0.04 \\
\midrule
TrustMix (ours) & 2.07 $\pm$ 0.02 & 2.72 $\pm$ 0.04 & \underline{2.21 $\pm$ 0.04} & 2.42 $\pm$ 0.05 \\
\bottomrule
\end{tabular}
\label{tab:results2}
\end{table}

\begin{table}[h]
\centering
\caption{95\% predictive interval coverage and width for all methods on UTK-Face. Displayed values are averages over 20 seeds with standard errors after $\pm$ sign. Best values and ties according to z-test in bold, second best are underlined.}
\begin{tabular}{c|cccc}
\toprule
Dataset & \multicolumn{4}{|c}{UTK} \\
Metric & ID Coverage  &  OOD Coverage &  ID Width ($\downarrow$) &  OOD Width ($\downarrow$) \\
\midrule
Base & \textbf{0.95 $\pm$ 0.01} & \underline{0.60 $\pm$ 0.05} & \underline{1.60 $\pm$ 0.11} &  \textbf{2.04 $\pm$ 0.46} \\
$\sigma$-scaling & \textbf{0.95 $\pm$ 0.00} & \underline{0.59 $\pm$ 0.04} & \underline{1.55 $\pm$ 0.06} &  \textbf{1.98 $\pm$ 0.45} \\
Dropout & \textbf{0.96 $\pm$ 0.01} & \underline{0.61 $\pm$ 0.04} & \underline{1.63 $\pm$ 0.11} & \textbf{2.06 $\pm$ 0.45} \\
$\sigma$-scaling + Dropout & \textbf{0.95 $\pm$ 0.00} & \underline{0.59 $\pm$ 0.04} & \underline{1.52 $\pm$ 0.06} & \textbf{1.93 $\pm$ 0.42} \\
Dist.-Aware $\sigma$-scaling & \textbf{0.95 $\pm$ 0.00} & \textbf{0.85 $\pm$ 0.02} & \underline{1.51 $\pm$ 0.05}  & 18.64 $\pm$ 5.83 \\
\midrule
TrustMix (ours) & \textbf{0.94 $\pm$ 0.00} & \textbf{0.83 $\pm$ 0.04} & \textbf{1.43 $\pm$ 0.03}  & \underline{2.70 $\pm$ 0.38} \\
\bottomrule
\end{tabular}
\label{tab:results3}
\end{table}

\newpage
\section{Further Results on Classification} \label{ap:class}
\begin{table}[H]
\centering
\caption{ECE ($\downarrow$) and NLL ($\downarrow$) on ID and OOD (corruption severity 5) datasets on CIFAR10 with the WideResNet 28-10 architecture. Reported values are means over 3 seeds, with values after $\pm$ being 1.96 times the standard error (half-width of the 95\% confidence interval). Best values and ties w.r.t. two-sided z-test are in bold, second best underlined.}
\footnotesize
\begin{tabular}{c|cc|cc}
\toprule
& \multicolumn{2}{c|}{$10 \times$ECE} & \multicolumn{2}{c}{NLL} \\
Method & ID & OOD & ID & OOD \\
\midrule
Base & 0.23 $\pm$ 0.01 & 2.50 $\pm$ 0.20 & 0.16 $\pm$ 0.00 & 1.78 $\pm$ 0.07 \\
DAC & \underline{0.09 $\pm$ 0.03} & \textbf{1.30 $\pm$ 0.26} & \underline{0.13 $\pm$ 0.00} & \textbf{1.34 $\pm$ 0.07} \\
HTS & 0.10 $\pm$ 0.01 & \underline{1.70 $\pm$ 0.22} & 0.14 $\pm$ 0.00 & \underline{1.44 $\pm$ 0.06} \\
IOP-IO & \textbf{0.04 $\pm$ 0.02} & \textbf{1.44 $\pm$ 0.19} & \textbf{0.13 $\pm$ 0.00} & \underline{1.47 $\pm$ 0.04} \\
IROVA & 0.11 $\pm$ 0.02 & \textbf{1.56 $\pm$ 0.23} & 0.20 $\pm$ 0.01 & 2.43 $\pm$ 0.55 \\
TS & 0.10 $\pm$ 0.02 & \underline{1.87 $\pm$ 0.17} & 0.14 $\pm$ 0.00 & \underline{1.47 $\pm$ 0.06} \\
\midrule
TrustMix (ours) & \underline{0.07 $\pm$ 0.00} & \textbf{1.30 $\pm$ 0.17} & \underline{0.13 $\pm$ 0.00} & \textbf{1.43 $\pm$ 0.07} \\
\bottomrule
\end{tabular}
\label{tab:table_cifar10_posthoc_wrn28_10}
\end{table}

\begin{table}[H]
\centering
\caption{ECE ($\downarrow$) and NLL ($\downarrow$) on ID and OOD (corruption severity 5) datasets on CIFAR10 with the VGG16BN architecture. Reported values are means over 3 seeds, with values after $\pm$ being 1.96 times the standard error (half-width of the 95\% confidence interval). Best values and ties w.r.t. two-sided z-test are in bold, second best underlined.}
\footnotesize
\begin{tabular}{c|cc|cc}
\toprule
& \multicolumn{2}{c|}{$10 \times$ECE} & \multicolumn{2}{c}{NLL} \\
Method & ID & OOD & ID & OOD \\
\midrule
Base & 0.32 $\pm$ 0.02 & 2.53 $\pm$ 0.22 & \underline{0.30 $\pm$ 0.01} & \underline{1.86 $\pm$ 0.11} \\
DAC & \textbf{0.09 $\pm$ 0.01} & \textbf{1.71 $\pm$ 0.21} & \textbf{0.28 $\pm$ 0.01} & \textbf{1.54 $\pm$ 0.06} \\
HTS & \textbf{0.09 $\pm$ 0.02} & \underline{1.90 $\pm$ 0.06} & \textbf{0.28 $\pm$ 0.01} & \textbf{1.58 $\pm$ 0.03} \\
IOP-IO & \textbf{0.11 $\pm$ 0.03} & \textbf{1.62 $\pm$ 0.19} & \textbf{0.28 $\pm$ 0.01} & \textbf{1.61 $\pm$ 0.13} \\
IROVA & \underline{0.13 $\pm$ 0.00} & \underline{1.91 $\pm$ 0.07} & 0.33 $\pm$ 0.02 & 2.18 $\pm$ 0.18 \\
TS & \textbf{0.09 $\pm$ 0.01} & \textbf{1.81 $\pm$ 0.11} & \textbf{0.28 $\pm$ 0.01} & \textbf{1.56 $\pm$ 0.04} \\
\midrule
TrustMix (ours) & \textbf{0.10 $\pm$ 0.01} & \textbf{1.64 $\pm$ 0.05} & \textbf{0.28 $\pm$ 0.01} & \textbf{1.58 $\pm$ 0.04} \\
\bottomrule
\end{tabular}
\label{tab:table_cifar10_posthoc_vgg16bn}
\end{table}

\begin{table}[H]
\centering
\caption{ECE ($\downarrow$) and NLL ($\downarrow$) on ID and OOD (corruption severity 5) datasets on CIFAR100 with the VGG16BN architecture. Reported values are means over 3 seeds, with values after $\pm$ being 1.96 times the standard error (half-width of the 95\% confidence interval). Best values and ties w.r.t. two-sided z-test are in bold, second best underlined.}
\footnotesize
\begin{tabular}{c|cc|cc}
\toprule
& \multicolumn{2}{c|}{$10 \times$ECE} & \multicolumn{2}{c}{NLL} \\
Method & ID & OOD & ID & OOD \\
\midrule
Base & 1.14 $\pm$ 0.03 & 3.32 $\pm$ 0.13 & \underline{1.23 $\pm$ 0.01} & 4.19 $\pm$ 0.13 \\
DAC & \textbf{0.24 $\pm$ 0.03} & \textbf{1.64 $\pm$ 0.24} & \textbf{1.12 $\pm$ 0.01} & \textbf{3.36 $\pm$ 0.09} \\
HTS & \underline{0.26 $\pm$ 0.02} & \underline{1.99 $\pm$ 0.15} & \textbf{1.12 $\pm$ 0.01} & \textbf{3.47 $\pm$ 0.07} \\
IOP-IO & \textbf{0.16 $\pm$ 0.09} & \textbf{1.90 $\pm$ 0.17} & \textbf{1.13 $\pm$ 0.01} & \underline{3.54 $\pm$ 0.07} \\
IROVA & 0.54 $\pm$ 0.05 & 2.29 $\pm$ 0.24 & 1.61 $\pm$ 0.05 & 7.16 $\pm$ 0.42 \\
TS & \textbf{0.24 $\pm$ 0.04} & \textbf{1.85 $\pm$ 0.17} & \textbf{1.13 $\pm$ 0.01} & \textbf{3.44 $\pm$ 0.08} \\
\midrule
TrustMix (ours) & \textbf{0.18 $\pm$ 0.03} & \textbf{1.73 $\pm$ 0.18} & \textbf{1.12 $\pm$ 0.01} & \textbf{3.48 $\pm$ 0.09} \\
\bottomrule
\end{tabular}
\label{tab:table_cifar100_posthoc_vgg16bn}
\end{table}

\begin{table}[H]
\centering
\caption{ECE ($\downarrow$) and NLL ($\downarrow$) on ID and OOD (corruption severity 5) datasets on ImageNet with the ResNet152 architecture. Reported values are means over 3 seeds, with values after $\pm$ being 1.96 times the standard error (half-width of the 95\% confidence interval). Best values and ties w.r.t. two-sided z-test are in bold, second best underlined.}
\footnotesize
\begin{tabular}{c|cc|cc}
\toprule
& \multicolumn{2}{c|}{$10 \times$ECE} & \multicolumn{2}{c}{NLL} \\
Method & ID & OOD & ID & OOD \\
\midrule
Base & 1.32 $\pm$ 0.01 & \textbf{0.40 $\pm$ 0.01} & 0.87 $\pm$ 0.01 & \textbf{3.49 $\pm$ 0.01} \\
DAC & 0.36 $\pm$ 0.02 & 1.32 $\pm$ 0.09 & \underline{0.73 $\pm$ 0.01} & \underline{3.51 $\pm$ 0.01} \\
HTS & 0.40 $\pm$ 0.00 & 2.05 $\pm$ 0.01 & \underline{0.73 $\pm$ 0.01} & 3.66 $\pm$ 0.03 \\
IOP-IO-Clip & \textbf{0.10 $\pm$ 0.04} & \underline{0.88 $\pm$ 0.09} & \textbf{0.67 $\pm$ 0.01} & \underline{3.55 $\pm$ 0.04} \\
IROVA & \underline{0.19 $\pm$ 0.02} & 1.89 $\pm$ 0.02 & 0.97 $\pm$ 0.03 & 10.84 $\pm$ 0.09 \\
TS & 0.35 $\pm$ 0.02 & 1.83 $\pm$ 0.01 & \underline{0.73 $\pm$ 0.01} & 3.61 $\pm$ 0.02 \\
\midrule
TrustMix (ours) & \underline{0.17 $\pm$ 0.03} & \underline{0.69 $\pm$ 0.23} & \textbf{0.68 $\pm$ 0.01} & 3.67 $\pm$ 0.01 \\
\bottomrule
\end{tabular}
\label{tab:table_imagenet_posthoc_resnet152}
\end{table}

\section{Detailed BO results} \label{ap:bo}

\begin{figure}[H]
    \centering
    \includegraphics[width=0.45\linewidth]{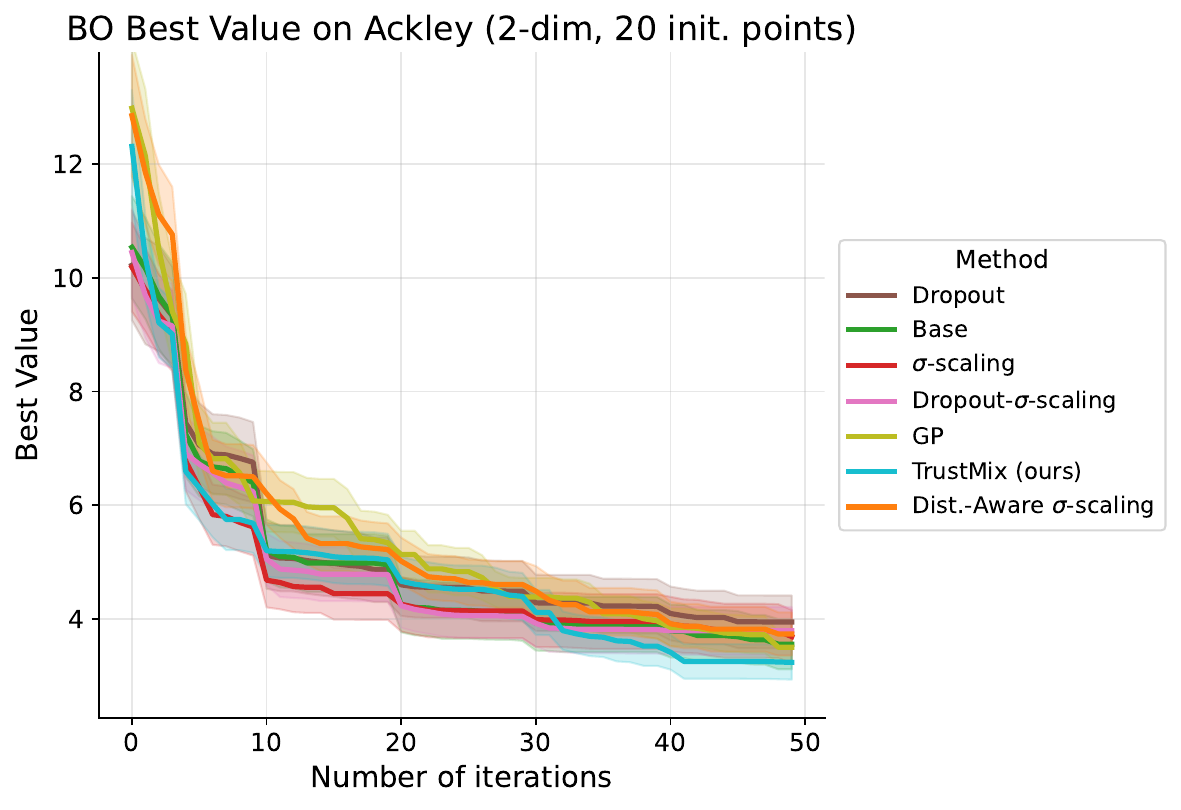}
    \includegraphics[width=0.45\linewidth]{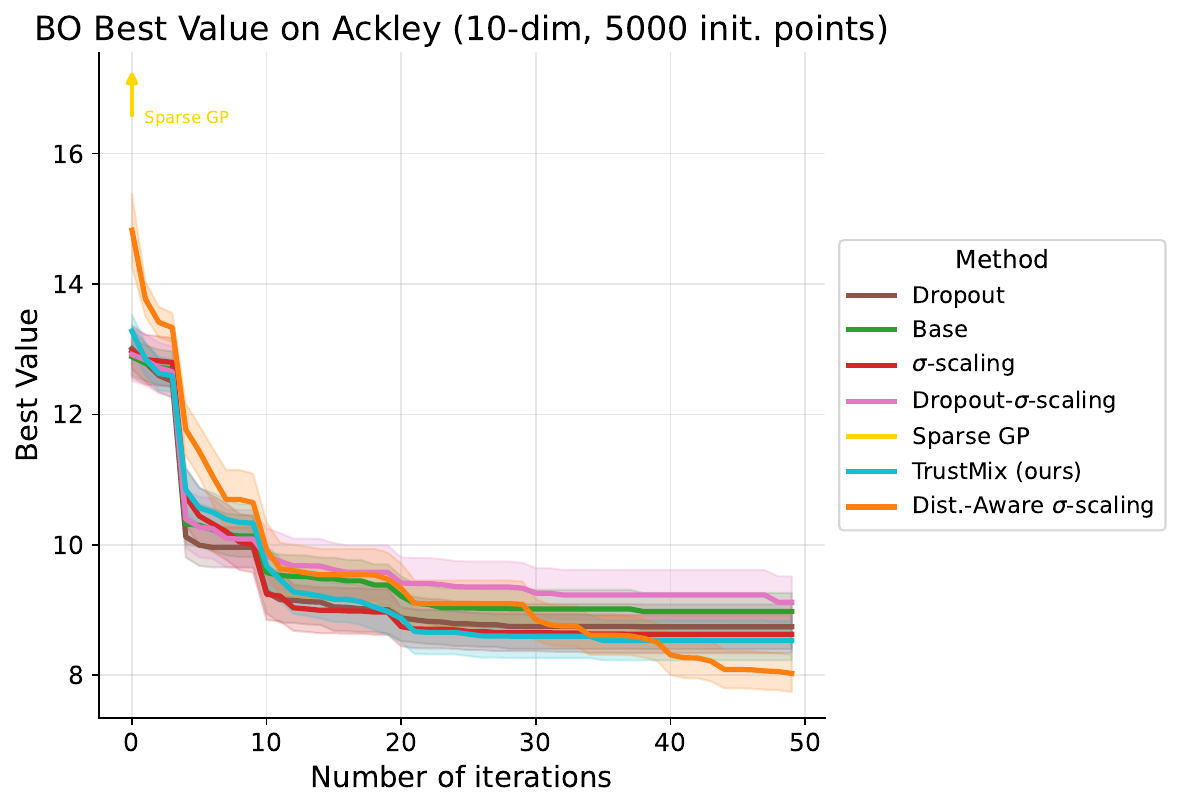}

    \includegraphics[width=0.45\linewidth]{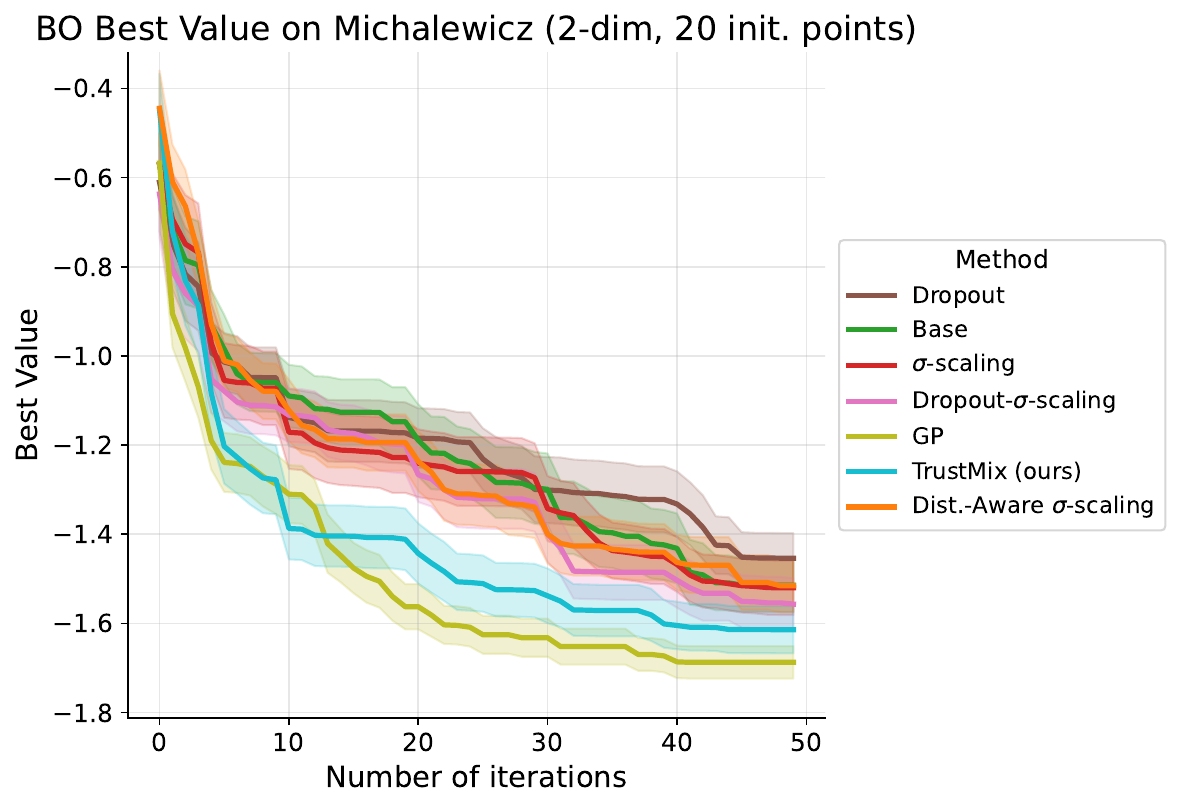}
    \includegraphics[width=0.45\linewidth]{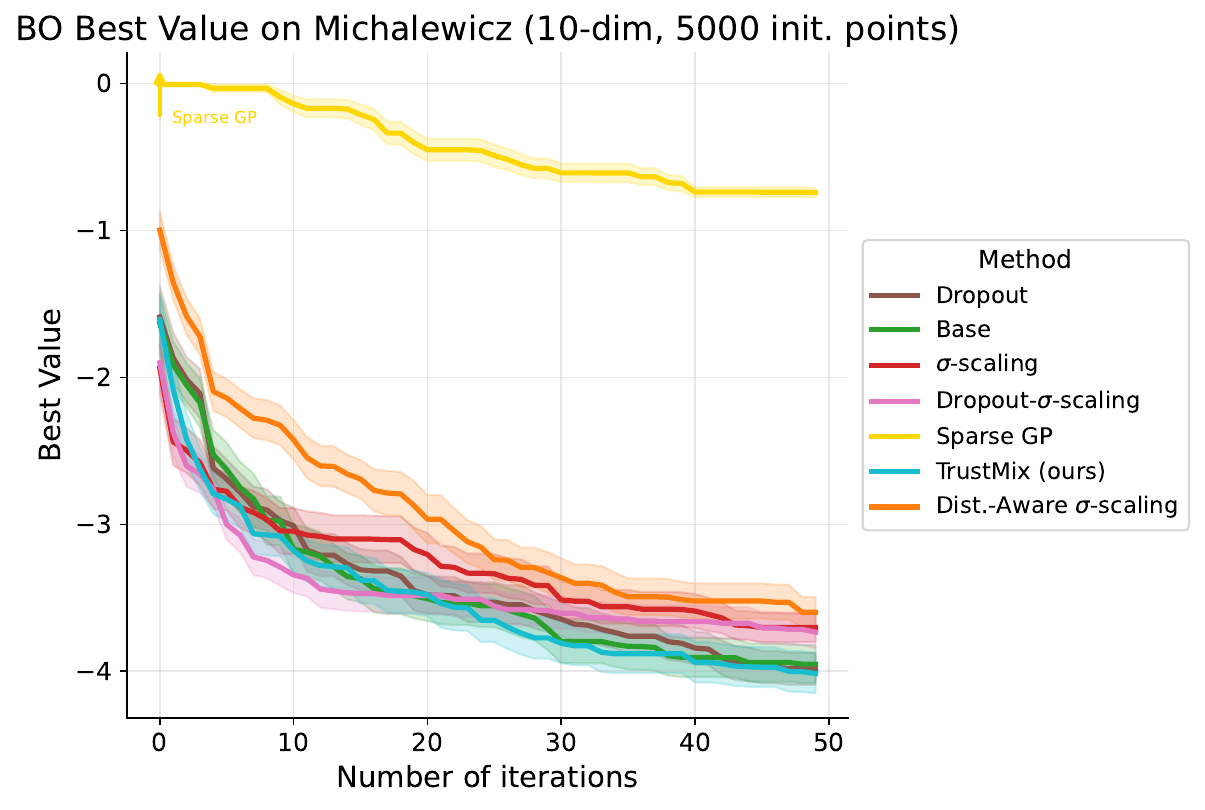}

\includegraphics[width=0.45\linewidth]{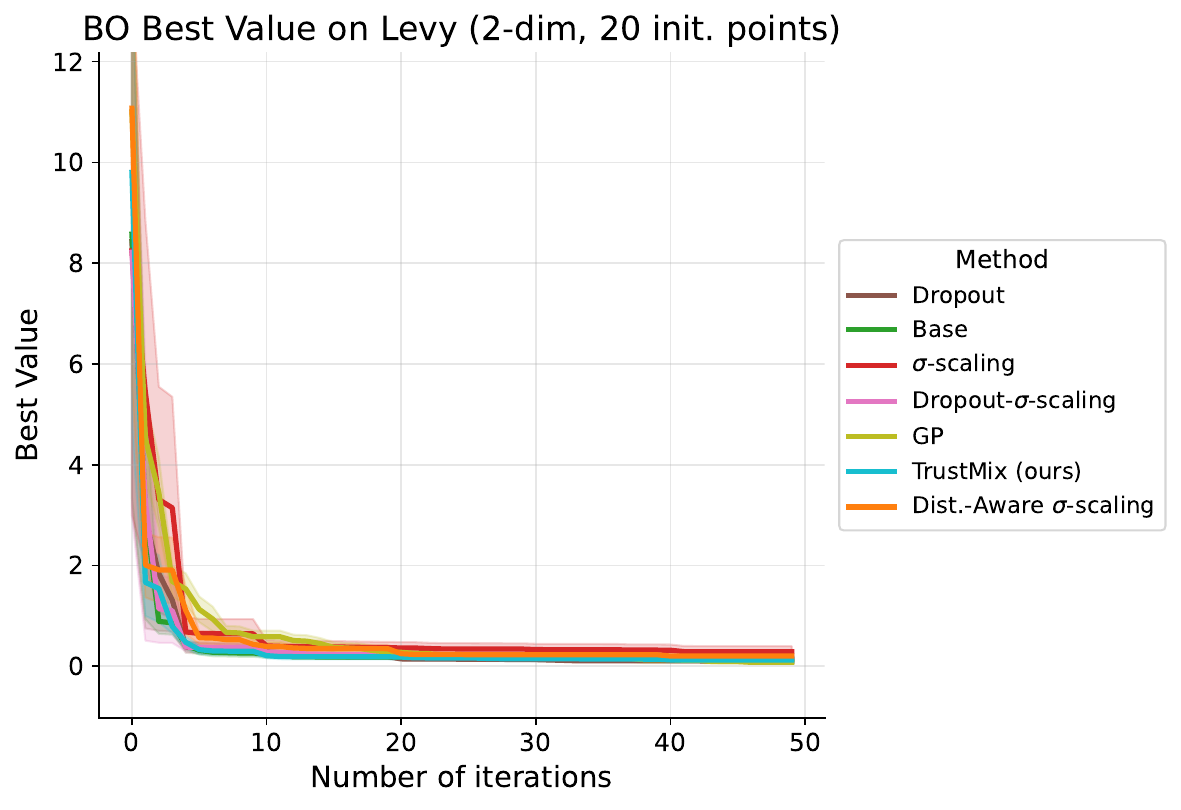}
    \includegraphics[width=0.45\linewidth]{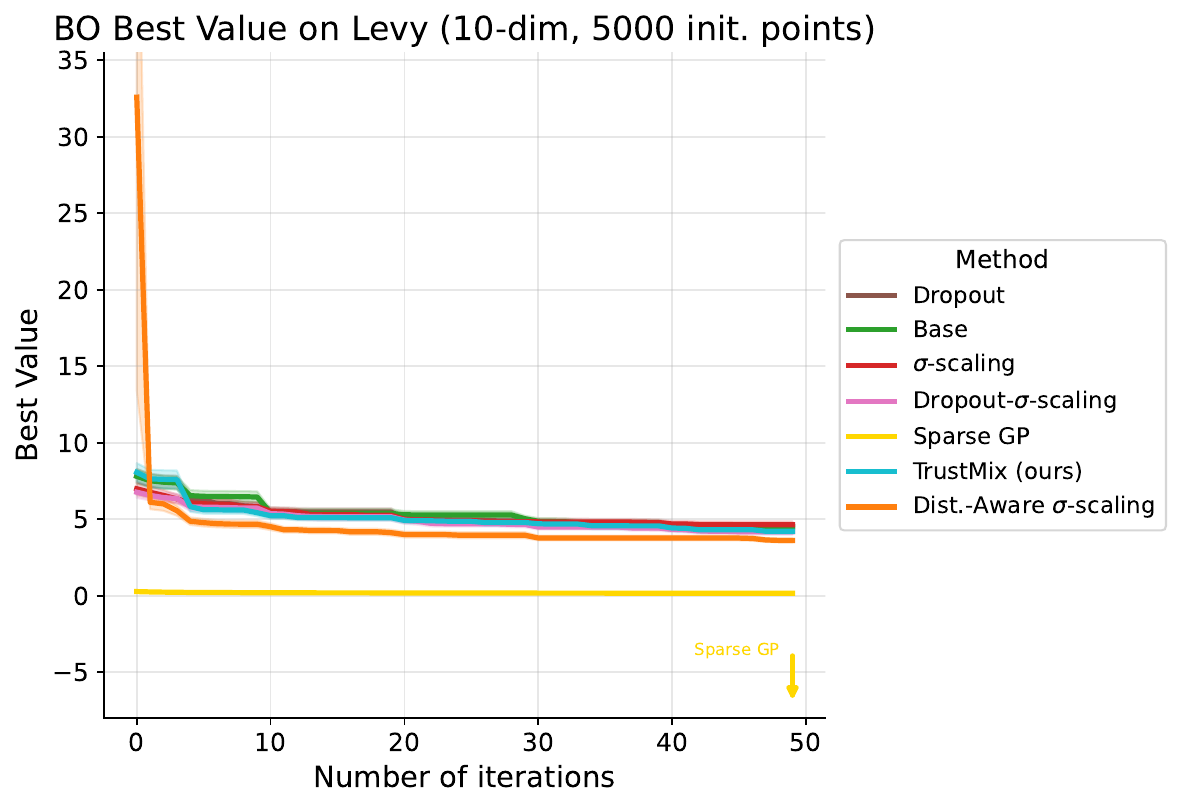}
\includegraphics[width=0.45\linewidth]{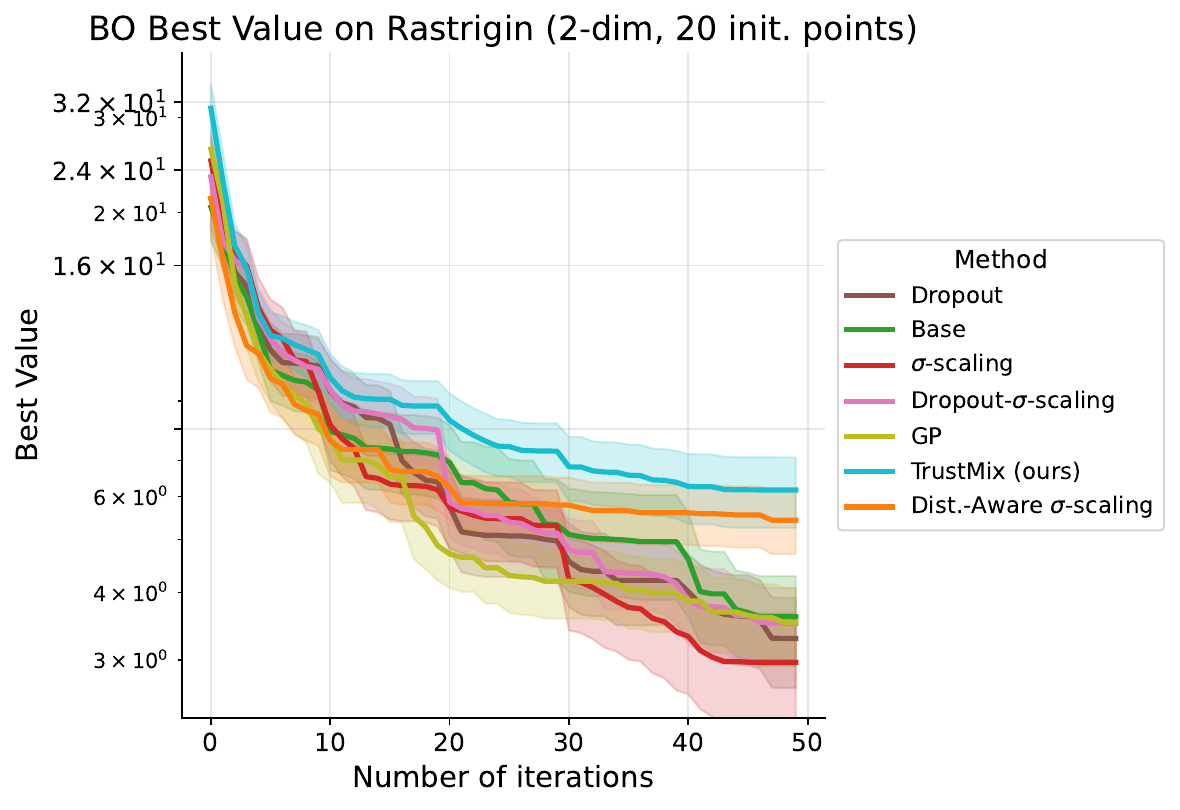}
    \includegraphics[width=0.45\linewidth]{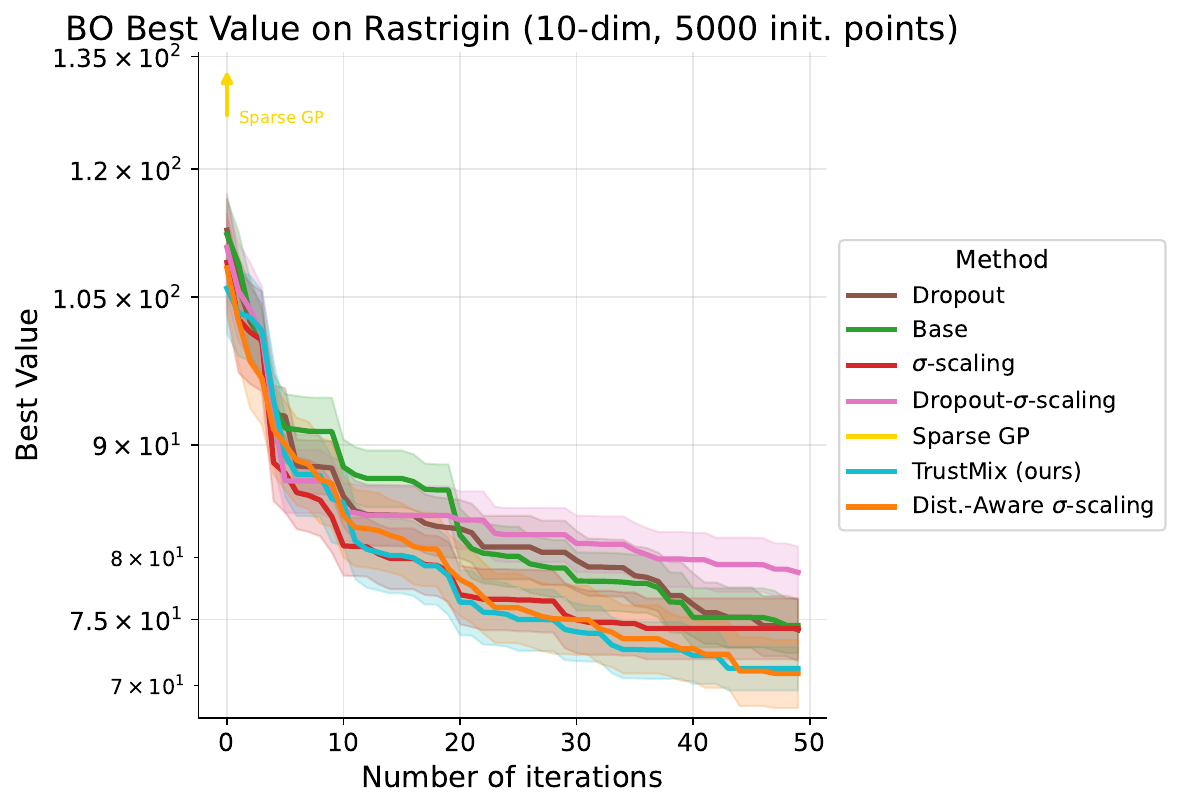}

    \caption{Detailed plots showing the average best value, with confidence intervals, achieved by each algorithm at each iteration (lower is better). Where a method performs substantially differently from the others, an arrow indicates that it lies outside the y-axis limits}
    \label{fig:detailed_bo}
\end{figure}

\begin{figure}
    \centering
    \includegraphics[width=0.45\linewidth]{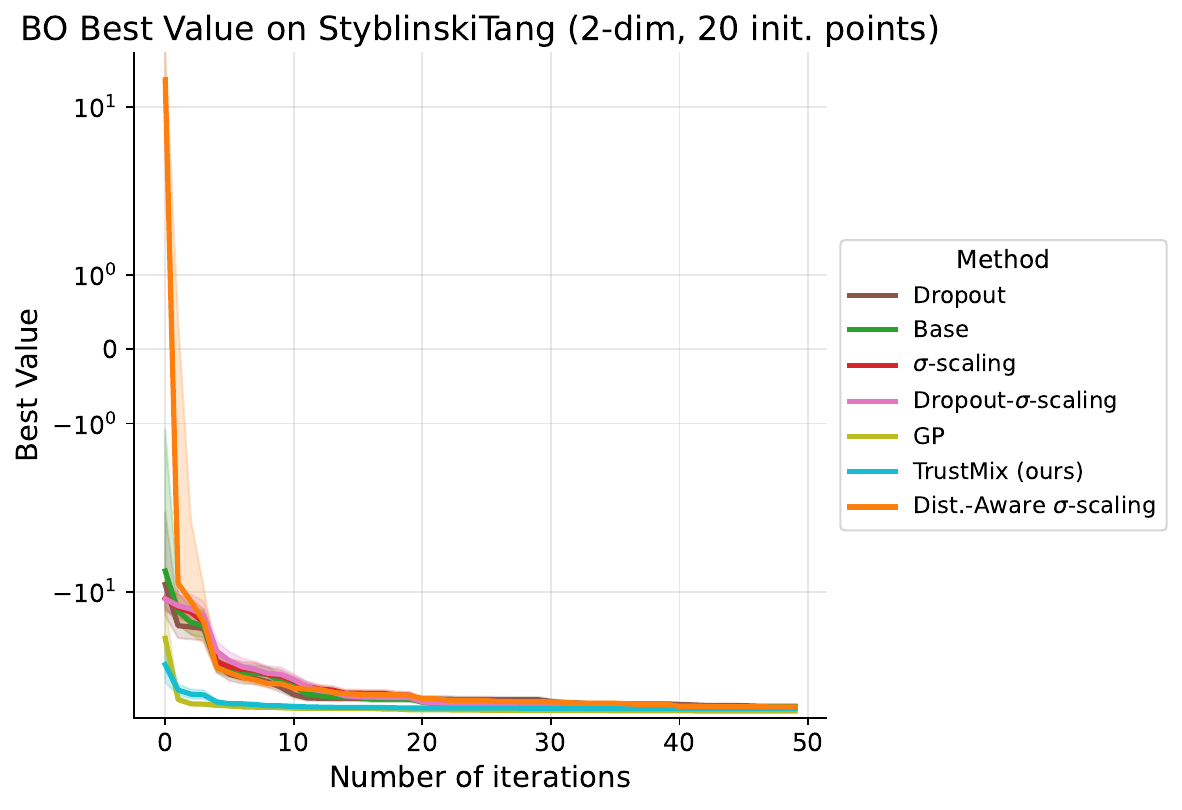}
    \includegraphics[width=0.45\linewidth]{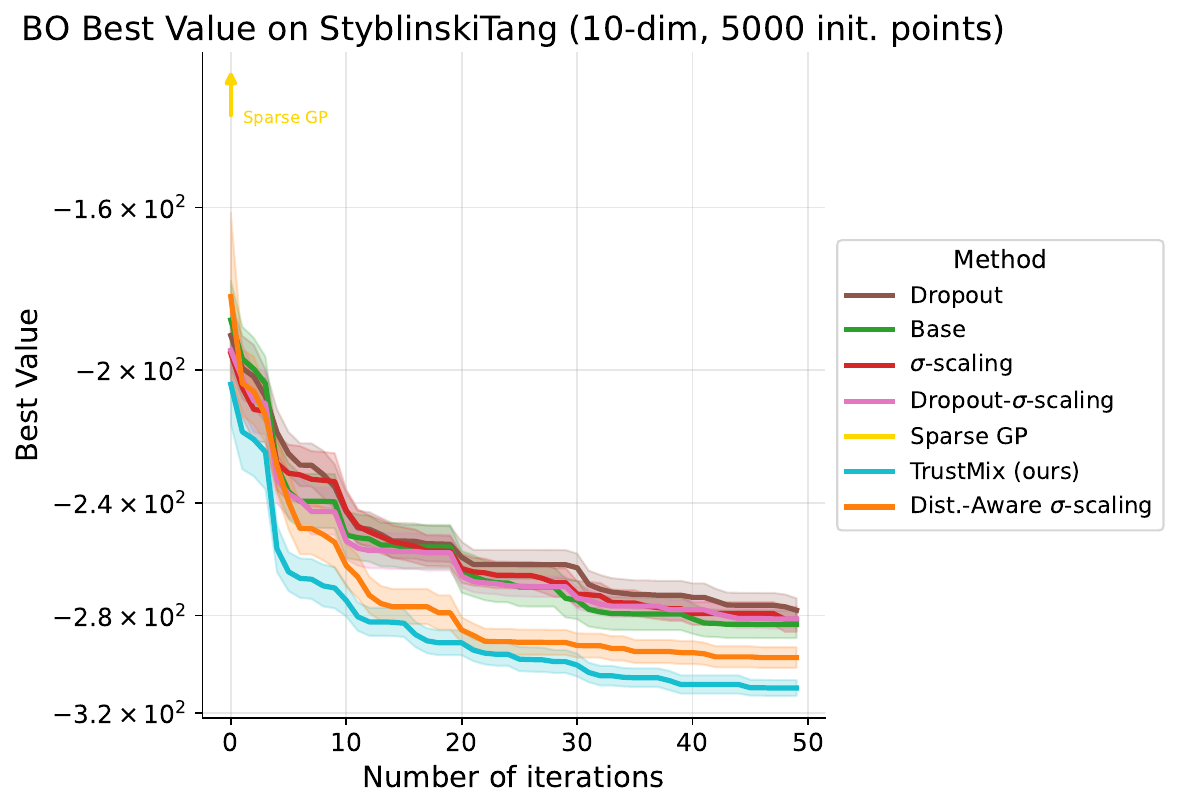}

\includegraphics[width=0.45\linewidth]{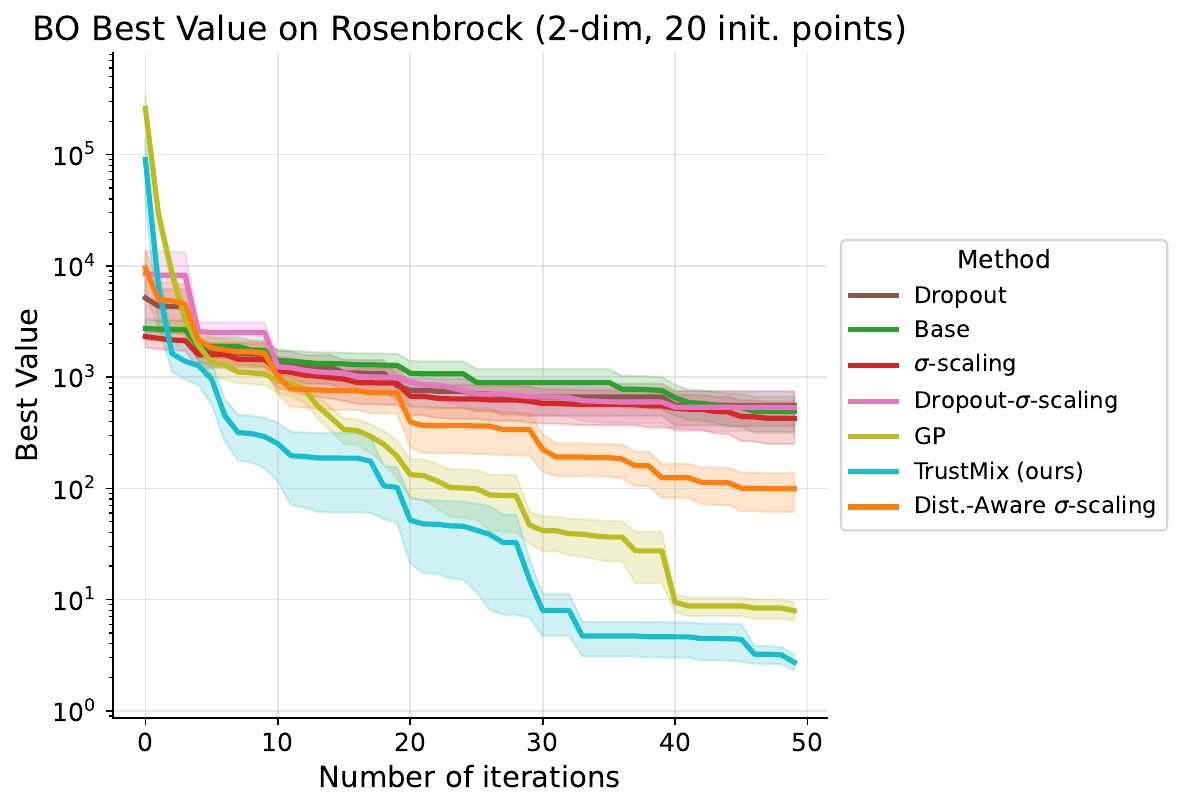}
    \includegraphics[width=0.45\linewidth]{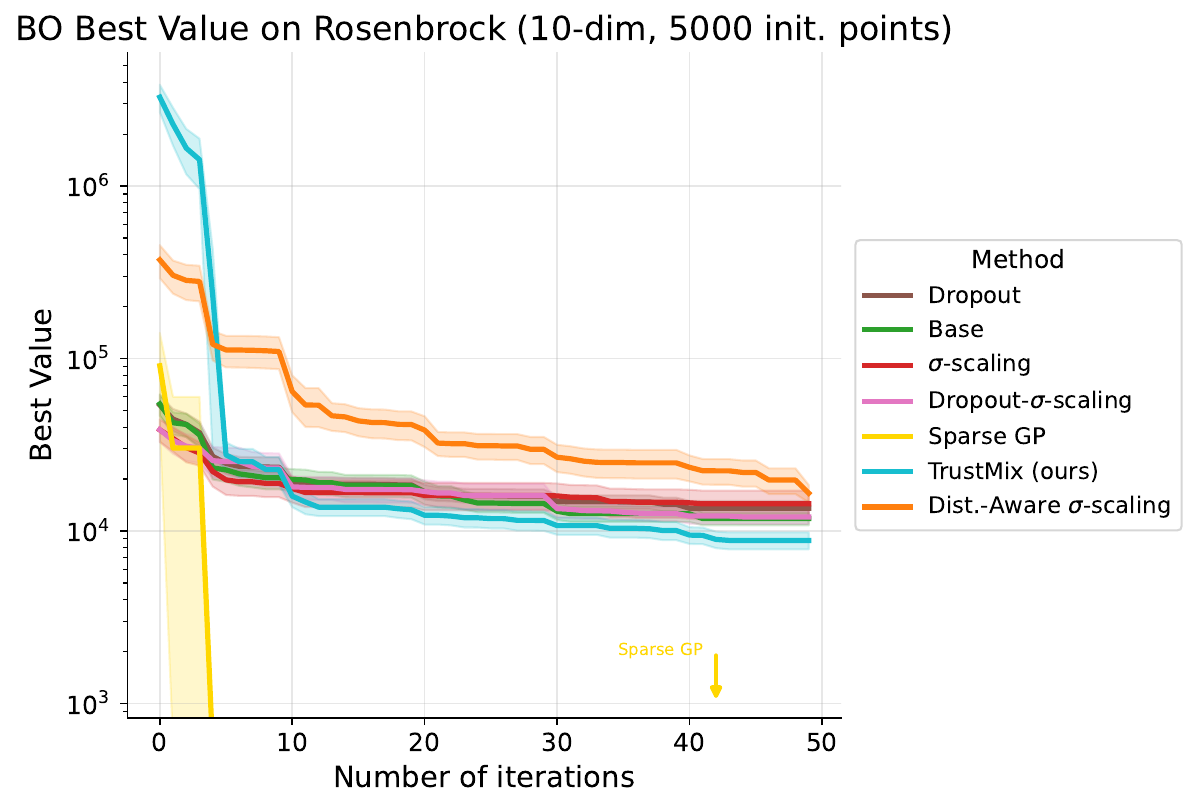}

    \includegraphics[width=0.45\linewidth]{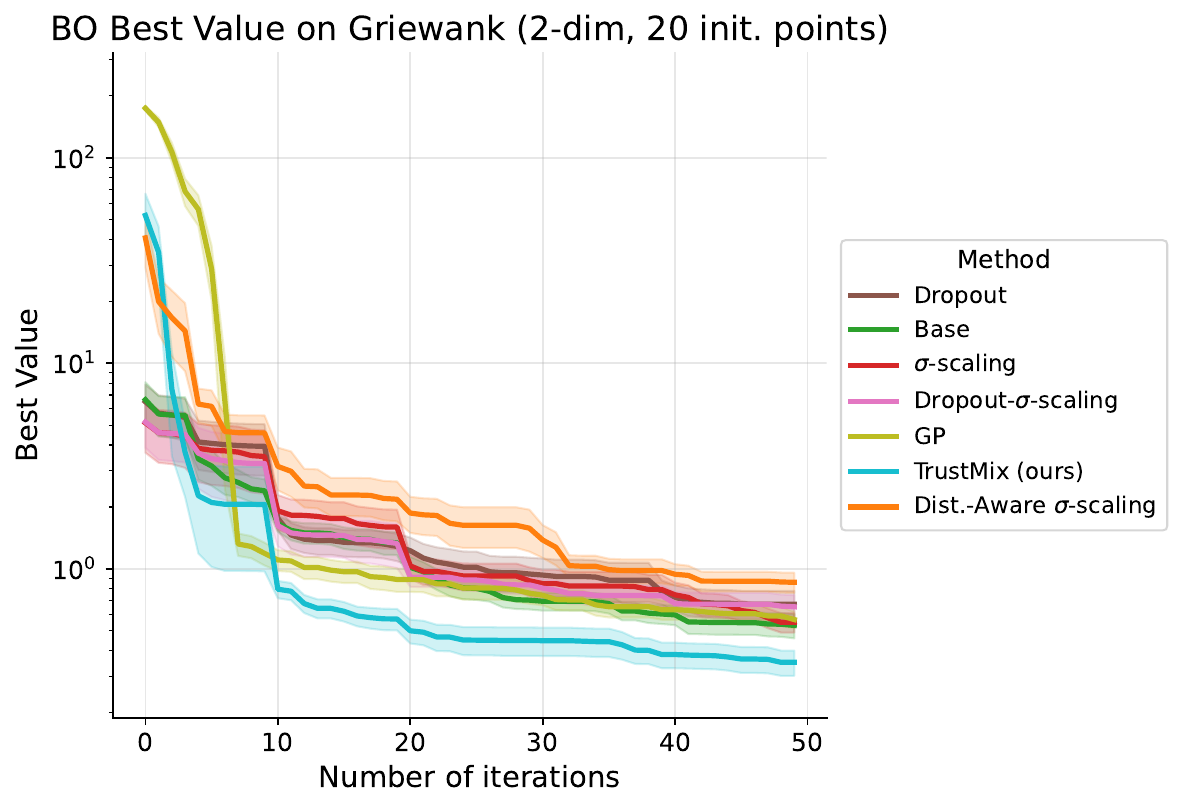}
    \includegraphics[width=0.45\linewidth]{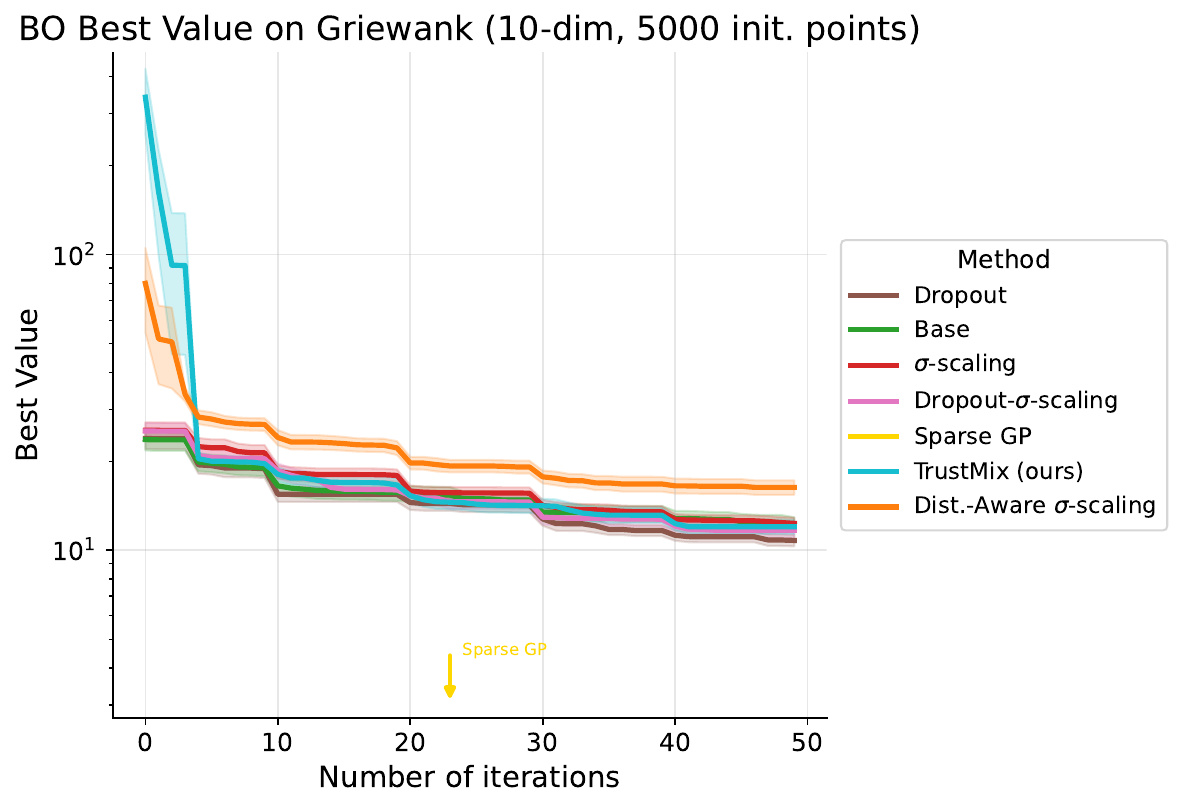}
    \caption{Detailed plots showing average based value with confidence intervals achieved by each algorithm by each iteration for each algorithm (lower the better). Where a method has vastly different performance to others, the arrow indicates that it lies outside of the y-axis limits.}
    \label{fig:detailed_bo2}
\end{figure}

\end{document}